\documentclass[lettersize,journal]{IEEEtran}
\usepackage{amsmath,amsfonts}
\usepackage{algorithmic}
\usepackage{algorithm}
\usepackage{array}
\usepackage[caption=false,font=normalsize,labelfont=sf,textfont=sf]{subfig}
\usepackage{textcomp}
\usepackage{stfloats}
\usepackage{url}
\usepackage{verbatim}
\usepackage{graphicx}
\usepackage{cite}
\usepackage{amsmath}
\usepackage{xcolor}
\usepackage{tcolorbox}
\tcbuselibrary{skins}

\usepackage{hyperref}
\usepackage{url}
\usepackage{multicol}
\allowdisplaybreaks
\usepackage{amssymb} 

\usepackage{booktabs}
\usepackage{multirow}  
\usepackage{array}  
\usepackage{makecell} 

\usepackage{wrapfig}
\usepackage{enumitem}
\usepackage[justification=centering]{caption}

\begin{document}

\title{A Unified Generalization Framework for Model Merging: Trade-offs, Non-Linearity, and Scaling Laws}







\author{Qinglun~Li,
        Anke~Tang,
        Miao~Zhang,
        Mengzhu~Wang,
        Quanjun~Yin,
        and~Li~Shen
\IEEEcompsocitemizethanks{
\IEEEcompsocthanksitem Q. Li and A. Tang contributed equally to this work.
\IEEEcompsocthanksitem M. Zhang and L. Shen are the corresponding authors. (E-mail: zhangmiao15@nudt.edu.cn; mathshenli@gmail.com)
\IEEEcompsocthanksitem Q. Li, M. Zhang, and Q. Yin are with the National University of Defense Technology, Changsha, China.
\IEEEcompsocthanksitem A. Tang is with Wuhan University, Wuhan, China.
\IEEEcompsocthanksitem L. Shen is with the School of Cyber Science and Technology, Shenzhen Campus of Sun Yat-sen University, Shenzhen, China.
\IEEEcompsocthanksitem M. Wang is with Hebei University of Technology, Tianjin, China.
}
}

        



\markboth{Journal of \LaTeX\ Class Files,~Vol.~14, No.~8, August~2021}%
{Shell \MakeLowercase{\textit{et al.}}: A Sample Article Using IEEEtran.cls for IEEE Journals}


\maketitle

\begin{abstract}
Model merging efficiently aggregates capabilities from multiple fine-tuned models into a single one, operating purely in parameter space without original data or expensive re-computation. Despite empirical successes, a unified theory for its effectiveness under heterogeneous finetuning hyperparameters (e.g., varying learning rates, batch sizes) remains missing. Existing federated learning theories focus purely on optimization, which fails to explain model merging and inherently leads to theoretical paradoxes. To address this challenge, we pioneer the integration of $L_2$-Stability theory into heterogeneous environments to rigorously decouple the excess risk of the merged model $\boldsymbol{x}_{avg}$ into optimization and generalization errors. This comprehensive analysis yields three main contributions: (i) We mathematically establish the fundamental \textit{Optimization-Generalization Trade-off}, explicitly resolving the paradox of why over-trained experts lead to catastrophic merging collapse. (ii) \textit{A unified theoretical framework} is provided to explain not only linear merging algorithms (e.g., TA, AdaMerging) but also state-of-the-art \textit{non-linear} merging algorithms (e.g., TIES, DARE),  proving how sparsification operators strictly tighten the generalization bound by suppressing task heterogeneity. (iii) Rather than heuristic guidelines, we derive \textit{Quantitative Scaling Laws} that theoretically predict the precise impact of hyperparameter choices, enabling practitioners to strategically construct ``merge-friendly'' experts. Extensive experiments on the ResNet and ViT architectures across 20 visual classification tasks, involving thousands of finetuning models, robustly confirm that our theoretical scaling laws accurately predict the empirical generalization behaviors of $\boldsymbol{x}_{avg}$.
\end{abstract}

\begin{IEEEkeywords}
Model Merging, Generalization Analysis, Convergence Analysis, Stability, Excess Error.
\end{IEEEkeywords}

\newtheorem{theorem}{Theorem}
\newtheorem{proposition}{Proposition}
\newtheorem{lemma}{Lemma}
\newtheorem{corollary}{Corollary}
\newtheorem{definition}{Definition}
\newtheorem{assumption}{Assumption}
\newtheorem{remark}{Remark}
\newtheorem{proof}{Proof}




\section{Introduction}
\label{sec:intro}

The paradigm of pre-training followed by fine-tuning has become the dominant paradigm in computer vision \cite{yuan2021florence,wortsmanRobustFinetuningZeroshot2022a}, leading to an explosion of powerful foundation models \cite{zheng2025learning,dosovitskiy2020image} and a vast ecosystem of specialized ``expert'' models fine-tuned for specific downstream tasks \cite{guo2019spottune}. Although these experts perform well on their respective tasks, selecting a specific expert model for each downstream task is cumbersome. A critical challenge has emerged: \textit{how can we efficiently aggregate the distinct capabilities of multiple expert models into a single, multi-talented model?} Model merging offers a compelling answer. By operating directly in the parameter space, methods like Model Soups~\cite{wortsman2022model,rameRewardedSoupsParetooptimal2023a,cheginiModelSoupBetter2024} and Task Arithmetic~\cite{TA2022editing} can fuse models without needing access to their original training data or incurring the prohibitive costs of re-training from scratch. Moreover, model merge techniques are also crucial for the advancement of current LLMs \cite{tam2024llm,fu2025training,wang2025optimal}.

\begin{figure}[t]
  \centering
  \includegraphics[width=0.48\textwidth]{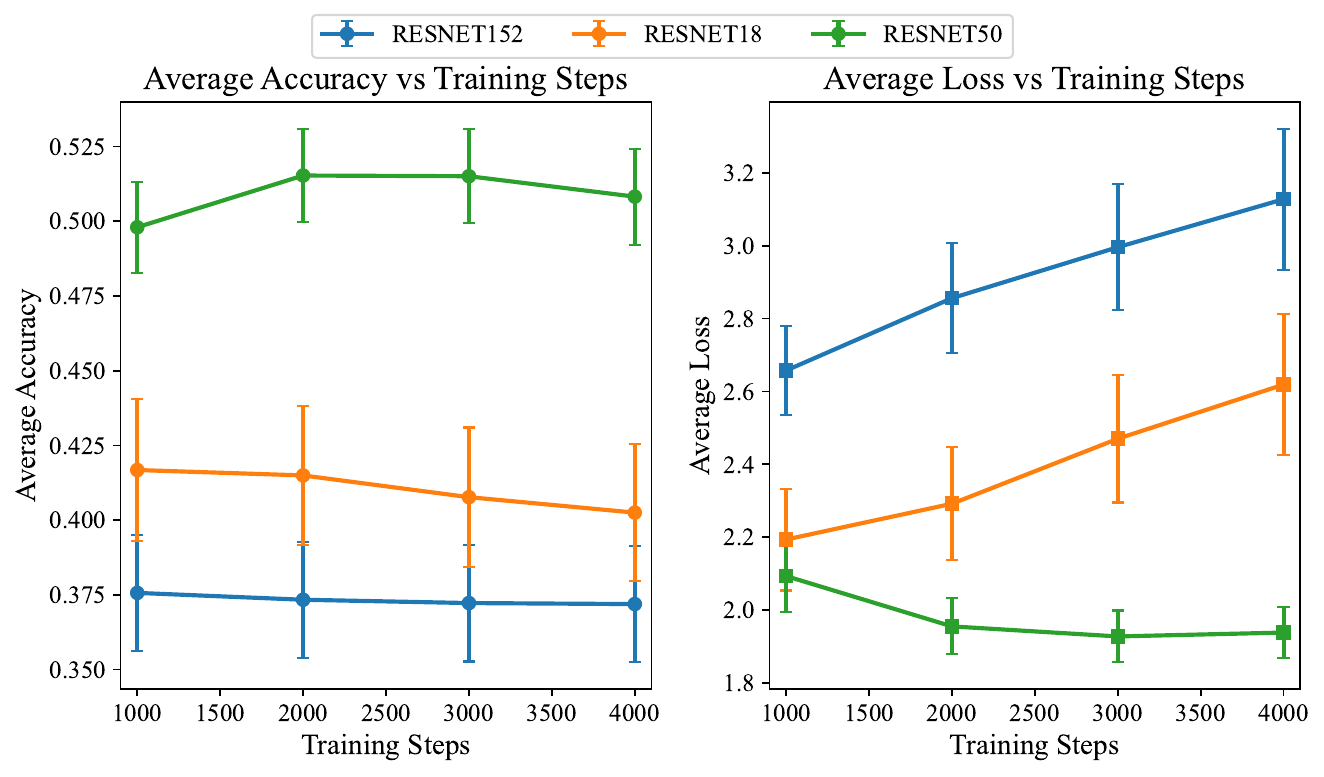}
\caption{\footnotesize{
  Using the Simple Average algorithm, we report Average Accuracy/Loss across different backbones. If only optimization theory is considered, one would conclude that the Average Loss decreases as $K$ increases. However, experimental results on ResNet-18 and ResNet-152 do not support this conclusion, indicating that optimization theory alone cannot explain the behavior of $\boldsymbol{x}_{{avg}}$.
  }}
  \label{fig:intro_figure}
  \vspace{-2em}
\end{figure}

Given its advantages, including operating directly in parameter space without requiring access to the original training data and its computational efficiency, researchers have proposed a wide range of effective techniques from different perspectives \cite{yadav2023ties, yu2024language, TA2022editing, yang2023adamerging, yang2024representation}. However, these methods are often explained heuristically, and there is still no unified theoretical framework to explain why and when they work. This theoretical gap also leads to significant practical challenges. For example, in open-source communities such as \textit{Hugging Face}, practitioners now have access to a large number of pretrained and fine-tuned models \cite{wolfHuggingFaceTransformersStateoftheart2020}. However, the key hyperparameters used during fine-tuning are often unavailable or non-transparent. This lack of transparency introduces two main difficulties: (i) The performance of model merging becomes highly unpredictable~\cite{yang2023adamerging}, and (ii) The trial-and-error process to identify suitable combinations greatly increases experimental cost~\cite{huangLoraHubEfficientCrossTask2024}. Furthermore, without theoretical guidance, practitioners attempting to merge multiple ``expert models'' face two central questions: (i) Since the underlying mechanisms that determine the effectiveness of different model merging methods remain unclear, it is difficult to select an appropriate merging strategy for a given set of models; and (ii) Without knowing how fine-tuning processes and their hyperparameters influence the final merging performance, it is challenging to fine-tune new models that can serve as good candidates for future merging—so-called \textit{merge-friendly} experts~\cite{wupiTuningTransferringMultimodal2023a,dziadzioHowMergeYour2025}.

Beyond the empirical challenges discussed above, a rigorous theoretical foundation remains elusive, particularly when aggregating experts fine-tuned under \textbf{heterogeneous hyperparameters} (e.g., diverse learning rates, batch sizes, and training steps). A seemingly intuitive approach is to borrow theoretical frameworks from Federated Learning (FL)~\cite{wang2020tackling}, which also aggregates decentralized models. However, mapping FL theories directly to model merging leads to a critical theoretical pitfall. Existing FL analyses primarily focus on the \textit{Optimization Error} ($\mathcal{E}_O$), asserting that a larger number of local fine-tuning steps ($K$) monotonically improves the aggregated model (typically $\mathcal{E}_O \propto 1/\sqrt{K}$) \cite{wang2020tackling}. Yet, our empirical observations in model merging fundamentally contradict this: over-trained experts often suffer from severe task interference, leading to a catastrophic collapse in the merged model's performance. 
As shown in Fig. \ref{fig:intro_figure}, results derived solely from optimization theory fail to explain the behavior of $\boldsymbol{x}_{\mathrm{avg}}$.
This contradiction indicates that optimization theory alone is fundamentally inadequate to explain the mechanics of model merging.


\begin{table*}[ht]
  \centering
  \caption{Summary of challenges in model merging and our contributions.}
  \label{tab:contributions}
  \resizebox{\linewidth}{!}{%
    \begin{tabular}{l|l|l}
      \toprule
      \textbf{Aspect}      & \textbf{Prevailing Challenges \& Gaps}                                & \textbf{Our Contribution}                          \\
      \midrule
      \textbf{Theoretical} & \begin{tabular}[c]{@{}l@{}}- Lack of a unified theory explaining both linear and non-linear methods.\\ - FL optimization theories fail in merging scenarios.\\ - Unclear why merging works under heterogeneous settings.\end{tabular} & \begin{tabular}[c]{@{}l@{}}\textbf{A Unified Generalization Framework}:\\ We derive the first excess error bound via $L_2$-Stability, resolving\\ the FL paradox and unifying linear/non-linear merging methods.\end{tabular} \\
      \midrule
      \textbf{Practical}   & \begin{tabular}[c]{@{}l@{}}- Merged model performance is highly unpredictable.\\ - Reliance on heuristic guidelines rather than exact formulas.\\ - High trial-and-error costs for practitioners.\end{tabular}           & \begin{tabular}[c]{@{}l@{}}\textbf{Quantitative Scaling Laws}:\\ Our bound provides precise mathematical formulas dictating\\ hyperparameter impact, enabling the creation of \textit{merge-friendly} experts.\end{tabular}          \\
      \midrule
      \textbf{Empirical}   & \begin{tabular}[c]{@{}l@{}}- Lack of systematic studies connecting theory to practice.\\ - Unclear if theoretical bounds generalize to modern architectures (ViTs)\\ and adaptive optimizers (Adam).\end{tabular}
      & \begin{tabular}[c]{@{}l@{}}\textbf{Extensive Empirical Validation}:\\ We evaluate thousands of models across ResNet and ViT architectures,\\ proving our theory aligns with empirical scaling trends.\end{tabular}             \\
      \bottomrule
    \end{tabular}}
  \vspace{-1em}
\end{table*}

To resolve this theoretical paradox and answer those practical questions, we argue that the success of model merging is not merely an optimization problem, but fundamentally a \textbf{generalization challenge}. In this paper, we pioneer the integration of $L_2$-Stability theory~\cite{lei2020fine,li2025unveiling} into the heterogeneous model merging environment. By rigorously decoupling the excess risk into Optimization Error ($\mathcal{E}_O$) and Generalization Error ($\mathcal{E}_G$), our framework yields four contributions:

\begin{itemize}[leftmargin=10pt]
    \item \textbf{Resolving the FL Paradox via an Optimization-Generalization Trade-off:} We formulate the first stability-based excess error bound for model merging. Crucially, our theory reveals that while increasing fine-tuning steps ($K$) or learning rates ($\eta_l$) reduces the optimization error, it simultaneously causes the stability penalty (and thus $\mathcal{E}_G$) to explode—mathematically scaled by $\mathcal{O}(K)$ and $\mathcal{O}(\eta_l^2)$. This rigorously establishes the \textit{Optimization-Generalization Trade-off}, successfully predicting the empirical ``inverted U-shape'' performance that previous theories failed to capture.
    
    \item \textbf{Theoretical Unification of Linear and Non-linear Merging Algorithms:} Beyond simple averaging~\cite{TA2022editing}, we mathematically extend our framework to explain state-of-the-art non-linear merging methods, such as TIES~\cite{yadav2023ties} and DARE~\cite{yu2024language}. We demonstrate how sparsity and magnitude-filtering operators strictly tighten our excess risk bound by explicitly suppressing the task heterogeneity term ($\zeta_i^2$).
    
    \item \textbf{Quantitative Scaling Laws for Merge-Friendly Experts:} Instead of heuristic guidelines, our bound translates empirical ``crafts'' into a predictive science by providing precise mathematical scaling laws. We quantitatively elucidate how hyperparameter choices (e.g., $\eta_l, b_i, K_i$) uniquely govern the merged model's generalizability, offering a strategic, theory-grounded guide for constructing merge-friendly experts.
    
    \item \textbf{Extensive Empirical Validation Across Architectures:} We validate our theory through thousands of fine-tuning and merging trials across diverse visual classification tasks~\cite{wangLocalizingTaskInformation2024a,2025Merging}. Moving beyond standard CNNs (ResNet) and SGD optimization, we extensively confirm that our theoretical scaling laws robustly generalize to Transformer architectures (ViT) optimized via adaptive methods (Adam), establishing a universal predictive framework.
\end{itemize}

Table~\ref{tab:contributions} summarizes the prevailing challenges and our corresponding theoretical and practical contributions.

\section{Related Works}

This section reviews diverse \textit{model merging methods} and the current \textit{theoretical advancement} in the field. At the end of each subsection, we explicitly outline the remaining gaps and open questions, thereby clearly positioning the contributions of this work.

\subsection{Model Merging Methods}
Model merging techniques can be broadly categorized by the stage at which they intervene as pre- and during-merging methods, following the taxonomy of \cite{yang2024model}.

\textbf{Pre-Merging Methods} focus on preparing models to be more compatible for merging, primarily by addressing the geometric misalignment of their parameter spaces.
A key strategy is \textit{Weight Alignment} \cite{ainsworth2023git,stoicaZipItMergingModels2023,xu2024weight,kindermanFoldableSuperNetsScalable2024,li2024training,naseryPLeaSMergingModels2025,jangModelStockAll2024,gargiuloTaskSingularVectors2025,marczakNoTaskLeft2025a}, which seeks to find permutation symmetries or other geometric relationships to align models within the same loss basin before averaging.
Git Re-basin~\cite{ainsworth2023git} and its extensions \cite{pena2023rebasin} exemplify this approach.
Another tactic is to modify the fine-tuning procedure itself, for instance, through \textit{Linearized Fine-tuning} in the tangent space of a pretrained model, which helps disentangle task-specific knowledge and reduce interference upon merging \cite{ortiz2023task,liuTangentModelComposition2023a}.

\textbf{During-Merging Methods} represent the core fusion algorithms. These methods span from basic methods to advanced strategies that mitigate task interference, representing one of the most active areas of current research.
These methods include:
(i) \textit{Basic methods:}
The simplest approaches include direct weight averaging, as popularized by Model Soups \cite{wortsman2022model,zimmerSparseModelSoups2023,chronopoulouAdapterSoupWeightAveraging2023a,rameRewardedSoupsParetooptimal2023a,biggsDiffusionSoupModel2024,zhouMetaGPTMergingLarge2024a,xieBoneSoupsSeekandSoup2025,kleimanSoupGoMitigating2025}, and Task Arithmetic \cite{TA2022editing,jinFineTuningAttentionModules2024,sunTaskArithmeticTrust2025}, which adds or subtracts vectors representing specific task capabilities. While foundational, their performance degrades when models exhibit conflicting parameter updates.
(ii) \textit{Weighted-based merging:}
To improve upon simple averaging, these methods assign adaptive weights to different models or their components. Some methods learn these coefficients using an auxiliary dataset \cite{yang2023adamerging}, while others rely on heuristics like the Fisher information matrix to determine parameter importance \cite{matena2022merging,crisostomiMASSMoErgingAdaptive2025,ostapenkoModularLLMsBuilding2024}.
(iii) \textit{Subspace-based merging:}
A highly effective strategy for reducing interference is first to sparsify models and merge them in a shared subspace~\cite{tangConcreteSubspaceLearning2023,yiSafetyRealignmentFramework2024a,heLocalizeandStitchEfficientModel2024a}.
TIES-Merging \cite{yadav2023ties} mitigates parameter conflicts by zeroing inconsistent signs and keeping high-magnitude updates, while DARE \cite{yu2024language} later showed that randomly pruning and rescaling task vectors can further enhance performance.
and (iv) \textit{Post-calibration methods:}
A more recent direction aims to calibrate the merged model after fusion.
This is motivated by the observation that even a successful merge in weight space can lead to a ``representation bias'', where the merged model's internal representations differ from those of the expert models \cite{yang2024representation,wanKnowledgeFusionLarge2023,wanFuseChatKnowledgeFusion2025}.
Representation Surgery \cite{yang2024representation} addresses this by introducing a lightweight module to correct the merged model's feature space in a post-hoc step.

\vspace{1mm}
\noindent \textbf{Open Questions:} Model merging has developed a rich set of empirically successful techniques, yet the underlying principles unifying their success remain unclear. Currently, no foundational theory explains, within a single framework, why diverse strategies—such as weight alignment before merging, sparsification during merging, or representation correction after merging—lead to improved merged models. This work addresses that gap by offering a unified theoretical perspective grounded in generalization theory.

\subsection{Theoretical Advancement in Model Merging}
Theoretical understanding of model merging significantly lags its empirical progress. As cataloged in \cite{yang2024model}, existing theoretical work can be grouped by the context in which models are merged.

First, a robust line of research analyzes the averaging of model checkpoints along a single training trajectory, including methods such as SWA~\cite{izmailov2018averaging,wang2024unified,wang2025sewa}, EMA~\cite{busbridgeHowScaleYour2023a,morales-brotonsExponentialMovingAverage2024}, LAWA~\cite{kaddourStopWastingMy2022a,ajroldiWhenWhereWhy2025a}, and SeWA~\cite{wangSeWASelectiveWeight2025}.
The success of these methods is generally attributed to their ability to converge to wide, flat minima in the loss landscape, which are known to promote better generalization~\cite{hardt2016train,wang2024generalization}. However, such analyses are confined to single-task settings and thus do not extend to our multi-task scenario.

Second, some studies explain the merging of models fine-tuned on the \textit{same dataset} but with different initializations or hyperparameters. The dominant theory here is \textit{Linear Mode Connectivity (LMC)} \cite{draxler2018essentially, entezari2022role,frankle2020linear}. 
LMC posits that the solutions found by SGD for overparameterized networks are not isolated points but are connected by linear paths of low loss. This provides a powerful geometric intuition for why models trained on the same data distribution can be effectively averaged, especially after alignment \cite{ainsworth2023git}.

Third, and most relevant to our paper, is the theory for merging models fine-tuned on \textit{different datasets or tasks}. This is the most challenging and least understood scenario. The few existing analyses are often tailored to specific methods. For example, \cite{ortiz2023task} use the Neural Tangent Kernel (NTK) to provide a compelling theoretical link between task arithmetic and the spectral properties of the model, but this analysis is specific to their proposed linearized fine-tuning method and is not applicable to heterogeneous hyperparameter settings of this paper.

\vspace{1mm}
\noindent \textbf{Open Questions:} The current theoretical landscape remains fragmented, lacking a unified framework that explains the generalization of merged models trained on diverse tasks under heterogeneous hyperparameter settings. Theories like LMC do not fully account for the data heterogeneity between tasks, and method-specific analyses do not provide a universal perspective. Therefore, a critical gap remains: We lack a theoretical framework that can both unify different merging methods and directly relate fine-tuning hyperparameters to the final generalization error of the merged model. This paper aims to fill this critical gap by leveraging $L_2$-Stability to derive the first such unifying generalization bound.
\section{Problem Formulation}

We consider $N$ tasks, each finetuned from a pretrained model with parameters $\boldsymbol{x}_0$. The goal of model merging is to independently train $N$ expert models on different tasks and find a merged parameter $\boldsymbol{x}_{avg}=\boldsymbol{x}_0 + \sum_{i=1}^N\lambda_i(\boldsymbol{x}_i - \boldsymbol{x}_0)$ that performs well across all $N$ tasks~\cite{TA2022editing,yang2023adamerging}, where $\boldsymbol{x}_i$ denotes the expert model parameters obtained by training $\boldsymbol{x}_0$ on task $i$. Specifically, model merging aims to minimize the following global population risk:
\begin{equation}
  F(\boldsymbol{x}) := \frac{1}{N} \sum_{i=1}^N \mathbb{E}_{z \sim \mathcal{P}_i} \left[ \ell(\boldsymbol{x}; z) \right],
\end{equation}
where $\boldsymbol{x}\in \mathbb{R}^d$ and $\mathcal{P}_i$ denotes the data distribution for task $i$, and the distributions differ across tasks. The loss function $\ell(\boldsymbol{x}; z): \mathcal{X} \times \mathcal{Z} \rightarrow \mathbb{R}^+$ is non-negative, and $z$ is sampled from the sample space $\mathcal{Z}$. Since the distributions $\mathcal{P}_i$ are typically unknown, the global population risk $F(\boldsymbol{x})$ cannot be computed exactly. Instead, we can access a training dataset $\mathcal{D} = \cup_{i=1}^N \mathcal{D}_i$, where $\mathcal{D}_i = \{ z_j^{(i)} \}_{j=1}^{n_i}$ is the training set for task $i$ of size $n_i$. We then approximate $F(\boldsymbol{x})$ using the global empirical risk:
\begin{equation}
  f(\boldsymbol{x}) := \frac{1}{N} \sum_{i=1}^N f_i(\boldsymbol{x}),
\end{equation}
where $f_i(\boldsymbol{x}) = \frac{1}{n_i} \sum_{j=1}^{n_i} \ell(\boldsymbol{x}, z_j^{(i)})$ is the empirical risk for task $i$, and the samples $z_j^{(i)} \overset{\text{i.i.d.}}{\sim} \mathcal{P}_i$ are drawn i.i.d. from $\mathcal{P}_i$.

To address the two questions raised in Section \ref{sec:intro}, we need to theoretically establish the relationship between the generalization bound of $\boldsymbol{x}_{avg}$ and the relevant hyperparameters. Next, we introduce several definitions for analyzing generalization bounds and explain their practical significance.

\textbf{Excess Error}\cite{lei2020fine,li2024boosting}. In this work, we focus on analyzing the excess error of algorithm, defined as $\mathcal{E}(\boldsymbol{x})\!:= \!F(\boldsymbol{x}) \!- \!f(\hat{\boldsymbol{x}})$,
where $\hat{\boldsymbol{x}}$ denotes the empirical risk minimizer (ERM). Moreover, the excess error can be decomposed as
\begin{equation}\label{eq:excess error}
  \mathcal{E}(\boldsymbol{x}) = \underbrace{F(\boldsymbol{x}) - f(\boldsymbol{x})}_{\mathcal{E}_G \text{ : generalization gap}} + \underbrace{f(\boldsymbol{x}) - f(\hat{\boldsymbol{x}})}_{\mathcal{E}_O \text{ : optimization error}}.
\end{equation}
Where the $\mathcal{E}_{G}$ represents the generalization error caused by approximating the unknown data distributions $\mathcal{P}_i,\forall i\in[N]$. The second term $\mathcal{E}_{O}$ captures the optimization error on the training data $\mathcal{D}$.

\begin{remark}
  \textbf{(Understanding Excess Error):} Excess error measures the gap between the risk on the true data distribution $F(\boldsymbol{x})$ and the empirical risk on the training dataset $f(\hat{\boldsymbol{x}})$. It can always be decomposed into the sum of the optimization error $\mathcal{E}_{O}$ and the generalization error $\mathcal{E}_{G}$. Specifically, the optimization error quantifies the gap between the solution $f(\boldsymbol{x})$ found by our algorithm on the training set and the theoretically optimal solution $f(\hat{\boldsymbol{x}})$, reflecting the capability and efficiency of the optimization algorithm itself. The generalization error measures the gap between the solution $f(\boldsymbol{x})$ found on the training set and the optimal solution $F(\boldsymbol{x})$ on the true data distribution, meaning how well the patterns learned from the training data generalize to unseen data. By combining both components, excess error provides a more comprehensive measure of an algorithm’s optimization efficiency and the quality of its solution.
\end{remark}

\begin{definition}\label{def:perturbed dataset}
  (Perturbed datasets). Let the global dataset be denoted by $\mathcal{D} = \bigcup_{i=1}^N\mathcal{D}_i$ and $\mathcal{D}_i = \{z_1, \dots, z_{n_i}\}$, where each $\mathcal{D}_i$ represents the local dataset of the $i$-th task with $|\mathcal{D}_i| = n_i$ for all $i \in [N]$. Consider another global dataset $\tilde{\mathcal{D}} = \bigcup_{i=1}^N \tilde{\mathcal{D}}_i$ and $\tilde{\mathcal{D}}_i= \{\tilde{z}_1, \dots, \tilde{z}_{n_i}\}$, independently sampled from $\mathcal{Z}$ such that $z_j, \tilde{z}_j \sim \mathcal{P}_i$ whenever $z_j \in \mathcal{D}_i$. For $j$, which is randomly selected from the indices $1, \dots, n_i$
  , define $\mathcal{D}^{(i)} \triangleq \{z_1, \dots, z_{j-1}, \tilde{z}_j, z_{j+1}, \dots, z_{n_i}\}$ as the perturbed version of $\mathcal{D}$, obtained by randomly replacing the $j$-th element with $\tilde{z}_j$, and define $\tilde{\mathcal{D}} \triangleq \bigcup_{j\neq i}\mathcal{D}_j\bigcup\mathcal{D}^{(i)}$.
\end{definition}

\begin{definition}[\textbf{$l_2$ on-average model stability}] \label{def:on-average-rephrased}
  Let $\mathcal{A}$ be a randomized algorithm. By leveraging the notation for perturbed datasets introduced in Definition \ref{def:perturbed dataset}, we say that $\mathcal{A}$ is \emph{$l_2$ on-average model $\varepsilon$-stable} if the following inequality holds:
  \begin{equation}
    \mathbb{E}_{\mathcal{D}, \tilde{\mathcal{D}}, \mathcal{A}}\left[\frac{1}{N} \sum_{i=1}^N \|\mathcal{A}(\mathcal{D}) - \mathcal{A}(\mathcal{D}^{(i)})\|_2^2\right] \leq \varepsilon^2 \;.
  \end{equation}
  The expectation $\mathbb{E}$ is taken over the random draws of the datasets $\mathcal{D}$ and $\tilde{\mathcal{D}}$, as well as the internal randomness of the algorithm $\mathcal{A}$.
\end{definition}

\begin{remark}
  \textbf{(Advantages of $L_2$-Stability):} Compared with Uniform Stability \cite{sun2021stability,liuunderstanding}, $L_2$-Stability \cite{lei2020fine,li2025unveiling} offers the following advantages: (1). Weaker assumptions and better practical applicability.
  Specifically, $L_2$-Stability removes the bounded-gradient assumption required by Uniform Stability. This represents a significant theoretical improvement, as in deep learning, there exist loss functions with unbounded gradients—for example, the mean squared error. Hence, the assumption behind $L_2$-Stability broadens its theoretical coverage and enhances its practical relevance. (2). Greater robustness to individual sample perturbations. Uniform Stability measures the worst-case upper bound, so if a single data point experiences a large perturbation, the bound becomes overly large. In contrast, $L_2$-Stability measures an average-case bound, making it more robust to variations in individual samples.
\end{remark}


\section{Theoretical Analysis}\label{sec:theoretical_analysis}

In this section, we first state the assumptions required for the theoretical analysis, then analyze the excess error $\mathcal{E}(\boldsymbol{x}_{avg})$ of model merge methods. Based on the derived excess-error bounds, we explain why different model merge methods are effective, how various hyperparameters influence their performance, and provide practical guidelines for improving empirical results.

\begin{assumption}\emph{($L$-smoothness).} \label{ass:smooth}The loss function $\ell$ is $L$-smooth i.e. $\exists L>0$ such that $\forall \boldsymbol{x}, \boldsymbol{y} \in \mathbb{R}^d, z \in \mathcal{D}$, $\|\nabla\ell(\boldsymbol{x};z)-\nabla\ell(\boldsymbol{y};z)\|_2\leq L \|\boldsymbol{x} - \boldsymbol{y}\|_2$.
\end{assumption}

\begin{assumption}\emph{(Bounded Variance).} \label{ass:bound SG} The local stochastic gradient $\boldsymbol{g}_i = \frac{1}{|\xi_i|} \sum_{z \in \xi_i} \nabla \ell(\boldsymbol{x}; z)$ is unbiased $\mathbb{E}[\boldsymbol{g}_i] = \nabla f_i(\boldsymbol{x})$ for all $\xi_i\in \mathcal{D}_i,i\in[N]$, Let $b_i = |\xi_i|$ denote the batch size used for task $i$ and there exists $\sigma_i^2>0$ such that $\mathbb{E}\|\boldsymbol{g}_i-\nabla f_i(\boldsymbol{x})\|^2\leq \frac{\sigma_i^2}{b_i}$, for all $\xi_i\in \mathcal{D}_i,i\in[N]$.
\end{assumption}

\begin{assumption}\emph{(Bounded Heterogeneity).}\label{ass:bound_hetero} There exists $\zeta_i^2>0$ such that $\mathbb{E}\|\nabla f_i(\boldsymbol{x})-\nabla f(\boldsymbol{x})\|^2\leq\zeta_i^2$, for any $i \in [N]$ and $\boldsymbol{x} \in \mathbb{R}^d$.
\end{assumption}

\begin{assumption}\emph{(Coefficients).}\label{ass:coeff}
  Let $\lambda_1, \dots, \lambda_N$ be the coefficients used to scale task vectors in the model merge algorithm. We assume that $\lambda_i \geq 0$ and $\sum_{i=1}^N \lambda_i = 1$.
\end{assumption}

\begin{assumption}\emph{(PL-Condition).}\label{ass:pl_condition}
There exists a constant $C > 0$ such that the optimization error is bounded by the gradient norm at the merged point: $\mathcal{E}_O(\boldsymbol{x}_{avg}) = f(\boldsymbol{x}) - f(\hat{\boldsymbol{x}}) \leq C \cdot \mathbb{E}\|\nabla f(\boldsymbol{x}_{avg})\|^2$.
\end{assumption}

\begin{remark}
  \textbf{(Rationale for Different Assumptions):}
  For Assumption \ref{ass:bound SG} and \ref{ass:bound_hetero}, unlike standard assumptions for algorithmic convergence, we adopt gradient variance and heterogeneity assumptions that are dependent on the index $i$, which aligns with the practical training hyperparameter settings of each expert model. For Assumption \ref{ass:pl_condition}, All expert models are fine-tuned from a high-quality, shared pretrained checkpoint $\boldsymbol{x}_0$. Recent literature \cite{Liu202285,neyshabur2020being} demonstrates that within such localized high-dimensional basins, PL-type conditions or mild weak-convexity hold as robust approximations.
\end{remark}

Notably, unlike analyses based on Uniform-Stability \cite{sun2021stability,liuunderstanding,li2024boosting}, our assumptions do not include the bounded-gradient assumption. Next, we present the generalization error bound and excess error bound based on $L_2$-Stability.

\subsection{Excess Error of Model Merging Method}

Before presenting the upper bound of Excess Error, we first introduce an important lemma that establishes the connection between $L_2$-Stability and Generalization Error. The proofs of all subsequent theorems and lemmas can be found in the Appendix~\ref{app:rheory}.

\begin{lemma}[\textbf{Generalization via on-average model stability}] \label{lemma:ob-avg-gen_main}
  Let $\mathcal{D}, \tilde{\mathcal{D}}, \mathcal{D}^{(i)}$ be constructed as Definition \ref{def:perturbed dataset}. Let $\gamma > 0$. If for any $z$, the function $f(\boldsymbol{x}; z)$ is nonnegative and $L$-smooth, then
  \begin{align*}
    \mathcal{E}_{G} & \leq \frac{1}{2\gamma} \mathbb{E}_{\mathcal{D}, \mathcal{A}}[\|\nabla f(\mathcal{A}(\mathcal{D}))\|^2] \\
    & + \frac{L + \gamma}{2} \frac{1}{N} \sum_{i=1}^N \mathbb{E}_{\mathcal{D}, \tilde{\mathcal{D}}, \mathcal{A}}[\|\mathcal{A}(\mathcal{D}^{(i)}) - \mathcal{A}(\mathcal{D})\|^2].
  \end{align*}
\end{lemma}
where $\mathcal{A}(\cdot)$ denotes a learning algorithm, and we denote $\boldsymbol{x} = \mathcal{A}(\mathcal{D})$ as the model produced by applying $\mathcal{A}$ to the dataset $\mathcal{D}$.
From Lemma \ref{lemma:ob-avg-gen_main}, we know that the generalization error $\mathcal{E}_{G}$ on the left-hand side is controlled by the upper bound of the gradient norm and $L_2$-Stability on the right-hand side. Therefore, we only need to derive the upper bound of the gradient norm and the $L_2$-Stability bound separately to control $\mathcal{E}_{G}$. Moreover, by combining a assumption \ref{ass:pl_condition} $\mathcal{E}_{O}(\boldsymbol{x}_{avg}) \leq C \cdot \mathbb{E}\|\nabla f(\boldsymbol{x}_{avg})\|^2$, we obtain the following stability-based expression for the excess error:
\begin{equation}\label{eq:excess_error_bound}
  \begin{aligned}
    \mathcal{E}(\boldsymbol{x}_{avg}) & \leq \frac{L + \gamma}{2} \underbrace{\mathbb{E}\|\boldsymbol{x}_{avg} - \tilde{\boldsymbol{x}}_{avg}\|^2}_{I_1:\textit{Model Stability}} \\&+ \left(\frac{1}{2\gamma} + C\right) \underbrace{\mathbb{E}\|\nabla f(\boldsymbol{x}_{avg})\|^2}_{I_2:\textit{Gradient Norm}}.
  \end{aligned}
\end{equation}
From inequality (\ref{eq:excess_error_bound}), we know that by controlling the upper bounds of the first term (\textit{Model Stability}) and the second term (\textit{Gradient Norm}) separately, we can obtain an upper bound on the excess error. In the following, we present the upper bounds for each term.

\textbf{(I). Upper Bound of $I_1$:} The proof of $I_1$ proceeds in two steps: first, derive $\tilde{I}_1 \triangleq \|\boldsymbol{x}_i^{K_i}-\tilde{\boldsymbol{x}}_i^{K_i}\|^2$; second, obtain an upper bound for $I_1$. Since $\boldsymbol{x}_{\textit{avg}}=\sum_{i=1}^N \lambda_i \boldsymbol{x}_i^{K_i},\tilde{\boldsymbol{x}}_{\textit{avg}}=\sum_{i=1}^N \lambda_i \tilde{\boldsymbol{x}}_i^{K_i}$ always holds, bounding $I_1$ requires first bounding $\tilde{I}_1$. The following theorem provides an upper bound for $I_1$.

\begin{theorem}\label{the:bound_of_stablity}
  \textbf{(Upper Bound of Model Stability):}
  Let $\mathcal{D},\ \tilde{\mathcal{D}}$ be constructed as in Definition $\ref{def:perturbed dataset}$. Let $\boldsymbol{x}_{\textit{avg}}$ and $\tilde{\boldsymbol{x}}_{\textit{avg}}$ denote the model parameters obtained by finetuning on $\mathcal{D}$ and $\tilde{\mathcal{D}}$, respectively, and then applying the model-merge method. Then we have the following:
  \begin{align*}
    \mathbb{E}_{\mathcal{D}, \tilde{\mathcal{D}}} \| \boldsymbol{x}_{avg} - \tilde{\boldsymbol{x}}_{avg} \|^2
    \leq 16\eta_l^2\sum_{i=1}^N\lambda_iK_i\left(\frac{\sigma_i^2}{n_i} + \frac{3b_i}{n_i}\zeta_i^2\right)
  \end{align*}
\end{theorem}

\textbf{(II). Upper Bound of $I_2$:} In real-world open-source model parameters, models $\boldsymbol{x}_i$ with different capabilities are often obtained using different training hyperparameters, such as different optimizers or varying numbers of training steps. Therefore, to make our results more consistent with practical scenarios, when bounding $I_2$, we introduce both \textit{training heterogeneity} (i.e., task-specific hyperparameters such as batch size $b_i$, number of steps $K_i$, gradient noise $\sigma_i$, etc., which differ across tasks) and \textit{data heterogeneity} (i.e., distributional divergence parameters $\zeta_i$ between tasks $i$ and $j$). This incorporation ensures that our results possess greater generality and broader applicability.

\begin{theorem}\label{the:Upper Bound of Gradient Norm}
  \textbf{(Upper Bound of Gradient Norm)}
  Under Assumptions \ref{ass:smooth}, \ref{ass:bound SG}, let $\bar{K} = \frac{1}{N} \sum_{i=1}^N K_i$, $\eta_l = \sqrt{\frac{N}{\bar{K}}}$ and all task use the SGD optimizer. Then, the gradient of the surrogate function $\tilde{f}(\boldsymbol{x}) = \sum_{i=1}^N \lambda_i f_i(\boldsymbol{x})$ at the averaged model $\boldsymbol{x}_{avg}$ is bounded as follows:
  \begin{equation}\label{eq:yizhi_upper bound of gradient norm}
    \begin{aligned}
       & \mathbb{E}[\|\nabla\tilde{f}(\boldsymbol{x}_{avg})\|^2] \leq \frac{4(\tilde{f}(\boldsymbol{x}_0) - \tilde{f}_{\inf})}{\sqrt{N\bar{K}}} +\frac{4L\sigma^2A_1}{b\sqrt{N\bar{K}}} \\
       & \hspace{3em} +\frac{6NL^2\sigma^2A_2}{b\bar{K}}+\frac{5NL^2\zeta^2A_3}{\bar{K}}.
    \end{aligned}
  \end{equation}
  Where $A_{1}=N\sum_{i=1}^{N}\frac{\bar{K}}{K_i}\lambda_{i}^{2}, A_2=\sum_{i=1}^N\lambda_i(K_i - 1), \tilde{f}_{\inf} = \underset{\boldsymbol{x}}{\min} \tilde{f}(\boldsymbol{x}), \sigma = \underset{i}{\max}{\{\sigma_i\}}, b =\underset{i}{\min}{ b_i},\zeta = \underset{i}{\max}{\{\zeta_i\}}$
  and $A_3 = \max_i\{K_i(K_i - 1)\}$.

  Furthermore, let the right-hand side of inequality (\ref{eq:yizhi_upper bound of gradient norm}) be denoted as $\epsilon_{sgd}$. Then, we have the following inequality, which ensures that even under heterogeneous conditions, the original objective function $f(\boldsymbol{x})$ can still converge, although its bound is not as tight as that in inequality \ref{eq:corollary1}:
  \begin{equation}\label{eq:upper bound of gradient norm}
    \mathbb{E}[\|\nabla f(\boldsymbol{x}_{avg})\|^2] \leq 2[\chi_{\boldsymbol{p}||\boldsymbol{\lambda}}^2 + 1]\epsilon_{sgd} + 2\chi_{\boldsymbol{p}||\boldsymbol{\lambda}}^2\sum_{i=1}^N\lambda_i\zeta_i^2
  \end{equation}
  where $\chi_{\boldsymbol{p}||\boldsymbol{\lambda}}^2 = \sum_{i=1}^N\frac{(\frac{1}{N} - \lambda_i)^2}{\lambda_i^2}$ is the chi-square divergence between the weight coefficient vectors $\boldsymbol{\lambda} =[\lambda_1,\lambda_2,\cdots,\lambda_N]\in \mathbb{R}^N$ and $\boldsymbol{p} =[\frac{1}{N},\frac{1}{N},\cdots,\frac{1}{N} ]\in \mathbb{R}^N$.
\end{theorem}

\textbf{(III). Upper Bound of Excess Error:}
By combining the bounds of $I_1$ and $I_2$ with inequality (\ref{eq:excess_error_bound}), we can readily obtain the upper bound of the excess error.

\begin{theorem}\label{the:tight upper bound of excess error}
  \textbf{(Tight Upper Bound of Excess Error)}
  By combining the conditions and results from Theorems \ref{the:bound_of_stablity} and \ref{the:Upper Bound of Gradient Norm}, and choosing an appropriate $\gamma^*$ in inequality (\ref{eq:excess_error_bound}), we obtain the following tighter excess error bound.
  \begin{align*}
    \label{eq:final bound of excess error}
    \mathcal{E}(\boldsymbol{x}_{avg}) \leq & 8(L+1)\eta_l^2\sum_{i=1}^N\lambda_iK_i\left(\frac{\sigma_i^2}{n_i} + \frac{3b_i}{n_i}\zeta_i^2\right)   \\
     & +(2C+1)\left[\chi_{\boldsymbol{p}||\boldsymbol{\lambda}}^2\sum_{i=1}^N\lambda_i\zeta_i^2 + (\chi_{\boldsymbol{p}||\boldsymbol{\lambda}}^2 + 1)\epsilon_{sgd}\right]
  \end{align*}
\end{theorem}

\subsection{Theoretical Unification of Non-Linear Merging Methods}
\label{sec:non_linear_unification}

A critical limitation of previous theoretical analyses is their confinement to linear weight combinations. State-of-the-art merging algorithms, such as TIES~\cite{yadav2023ties} and DARE~\cite{yu2024language}, employ non-linear transformations (e.g., magnitude pruning, random dropping, and sign-consensus) prior to merging. In this section, we extend our framework to demonstrate how non-linear operators fundamentally tighten our $L_2$-stability excess risk bound.

In the parameter space, the task-specific update $\Delta \boldsymbol{x}_i = \boldsymbol{x}_i - \boldsymbol{x}_0$ serves as a surrogate for the accumulated negative gradient direction. Under Assumption \ref{ass:bound_hetero}, the task heterogeneity $\zeta_i^2$ is inherently bounded by the variance of these update directions across different tasks, i.e., the presence of ``parameter interference'' or conflicting signs. 

We define a general \textbf{Sparsification and Consensus Operator} $\mathcal{M}(\cdot): \mathbb{R}^d \rightarrow \mathbb{R}^d$, which encompasses both TIES (retaining top-$k\%$ magnitudes and resolving sign conflicts) and DARE (random dropping with rescaling). Let $\Delta \tilde{\boldsymbol{x}}_i = \mathcal{M}(\Delta \boldsymbol{x}_i)$ be the modified task vector. The non-linear merged model is thus formulated as $\tilde{\boldsymbol{x}}_{avg} = \boldsymbol{x}_0 + \sum_{i=1}^N \lambda_i \mathcal{M}(\Delta \boldsymbol{x}_i)$.

\begin{corollary}\label{cor:non_linear}
  \textbf{(Excess Risk Reduction via Non-Linear Sparsification)}
  Suppose the operator $\mathcal{M}(\cdot)$ filters out parameter updates that possess conflicting signs across tasks (sign-consensus) or fall below a magnitude threshold $\tau$ (pruning). By zeroing out these interference components, the empirical variance among task gradients is strictly reduced. Consequently, the effective task heterogeneity $\hat{\zeta}_i^2$ for the modified updates satisfies:
  \begin{equation}
      \hat{\zeta}_i^2 = \zeta_i^2 - \delta_i(\mathcal{M}), \quad \text{where } \delta_i(\mathcal{M}) > 0
  \end{equation}
  Here, $\delta_i(\mathcal{M})$ represents the variance reduction strictly yielded by eliminating conflicting and low-magnitude updates. Substituting $\hat{\zeta}_i^2$ into the excess risk bound $\mathcal{E}(\boldsymbol{x}_{avg})$ from Theorem \ref{the:tight upper bound of excess error}, the new excess risk bound for the non-linearly merged model $\mathcal{E}(\tilde{\boldsymbol{x}}_{avg})$ satisfies:
  \begin{equation}
      \mathcal{E}(\tilde{\boldsymbol{x}}_{avg}) < \mathcal{E}(\boldsymbol{x}_{avg})
  \end{equation}
\end{corollary}

\begin{proof}
Recall that under Assumption \ref{ass:bound_hetero}, the local gradient variance $\zeta_i^2$ bounds the expected deviation: $\mathbb{E}\|\nabla f_i(\boldsymbol{x}) - \nabla f(\boldsymbol{x})\|^2 \leq \zeta_i^2$. In the parameter space, the task vector $\Delta \boldsymbol{x}_i$ corresponds to the accumulated task-specific gradients. The heterogeneity $\zeta_i^2$ is thus empirically proportional to the variance of these task vectors: $\mathbb{E}\|\Delta \boldsymbol{x}_i - \Delta \bar{\boldsymbol{x}}\|^2$, where $\Delta \bar{\boldsymbol{x}} = \sum_{k=1}^N \lambda_k \Delta \boldsymbol{x}_k$ is the consensus mean.

The non-linear operator $\mathcal{M}(\cdot)$ applies a task-specific sparse mask $\boldsymbol{m}_i \in \{0,1\}^d$ such that $\Delta \tilde{\boldsymbol{x}}_i = \boldsymbol{m}_i \odot \Delta \boldsymbol{x}_i$. Specifically, methods like TIES enforce \textit{sign consensus}, meaning $m_{i,j} = 0$ if the sign of the $j$-th coordinate $\Delta x_{i,j}$ contradicts the dominant direction of $\Delta \bar{x}_j$.

For any dimension $j$, the original variance contribution is $V_{i,j} = (\Delta x_{i,j} - \Delta \bar{x}_j)^2$. If $\Delta x_{i,j}$ is a conflicting update (e.g., $\Delta x_{i,j} < 0$ while the consensus $\Delta \bar{x}_j > 0$), it severely inflates the distance $V_{i,j}$. By applying the operator $\mathcal{M}$, this conflicting component is zeroed out ($\Delta \tilde{x}_{i,j} = 0$). Since $0$ is strictly closer to the consensus mean $\Delta \tilde{\bar{x}}_j$ than the original conflicting value $\Delta x_{i,j}$, the deviation strictly decreases:
\begin{equation*}
    (0 - \Delta \tilde{\bar{x}}_j)^2 < (\Delta x_{i,j} - \Delta \bar{x}_j)^2
\end{equation*}
Let $\mathcal{S}_{drop}$ denote the set of dimensions pruned by $\mathcal{M}$. The exact reduction in the heterogeneity bound for task $i$ can be quantified as:
\begin{equation}
    \delta_i(\mathcal{M}) \propto \sum_{j \in \mathcal{S}_{drop}} \left[ (\Delta x_{i,j} - \Delta \bar{x}_j)^2 - (0 - \Delta \tilde{\bar{x}}_j)^2 \right] > 0
\end{equation}
Because the operator explicitly removes outlier coordinates that inflate variance, the effective heterogeneity for the modified updates strictly satisfies $\hat{\zeta}_i^2 = \zeta_i^2 - \delta_i(\mathcal{M})$. 

According to Theorem \ref{the:tight upper bound of excess error}, the excess risk bound $\mathcal{E}(\boldsymbol{x}_{avg})$ is a monotonically increasing function of $\zeta_i^2$. The $\zeta_i^2$ term appears with positive coefficients in both the \textit{Model Stability} penalty (scaled by $\frac{3b_i}{n_i}$) and the \textit{Optimization Error} term (scaled by $\chi_{\boldsymbol{p}||\boldsymbol{\lambda}}^2$). Substituting $\hat{\zeta}_i^2 < \zeta_i^2$ into these terms trivially yields $\mathcal{E}(\tilde{\boldsymbol{x}}_{avg}) < \mathcal{E}(\boldsymbol{x}_{avg})$, completing the proof.
\end{proof}

\begin{remark}\label{remark:different method of interpretation}
  \textbf{(Interpretation of Classical Model Merge Methods via the Excess Error Bound):}
  The excess error bound presented in Theorem \ref{the:tight upper bound of excess error} and Corollary \ref{cor:non_linear} provides a unified theoretical framework for understanding the mechanisms behind various model merging methods. Each method can be interpreted as modifying different terms in our bounds to balance the optimization error $\mathcal{E}_{O}$ and the generalization error $\mathcal{E}_{G}$.
  \looseness =-1
  \begin{itemize}[leftmargin=5pt]
    \item \textbf{Pre-Merging Methods (e.g., Git Re-basin \cite{ainsworth2023git}):} These methods focus on aligning the parameter spaces of expert models before they are averaged. From our theoretical perspective, this alignment serves to reduce the initial task heterogeneity $\zeta_i^2$. By finding permutation symmetries that place models in a common basin, these techniques ensure that the subsequent merging step operates on a set of models that are already more geometrically and functionally concordant. Moreover, \cite{lee2025mitigating} (SAM) minimizes the dominant eigenvalue of the Hessian ($\lambda_{max}(\mathbf{H})$), which effectively reduces the local Lipschitz constant $L$, thereby directly shrinking the stability penalty. Similarly, \cite{tang2023parameter} (Linearization) enforces near-zero curvature (Hessian $\approx 0$), minimizing non-linear deviations and strictly tightening the stability bound.

    \item \textbf{During-Merging Methods:} These algorithms directly manipulate the fusion process itself. 
    (i) \textit{Basic Methods} (Simple Averaging \cite{wortsman2022model}, Task Arithmetic \cite{TA2022editing}): This approach sets $\lambda_i=1/N$, greedily eliminating the chi-square divergence penalty $\chi_{\boldsymbol{p}||\boldsymbol{\lambda}}^2$. However, its agnosticism to task heterogeneity makes it vulnerable to outlier tasks with large $\zeta_i^2$. 
    (ii) \textit{Weighted-based Merging} (e.g., AdaMerging \cite{yang2023adamerging}): They learn non-uniform coefficients $\boldsymbol{\lambda}$ that strategically accept a non-zero surrogate penalty ($\chi_{\boldsymbol{p}||\boldsymbol{\lambda}}^2 > 0$) in exchange for a greater reduction in the heterogeneity-dependent terms ($\sum \lambda_i \zeta_i^2$), explicitly down-weighting outlier tasks. 
    (iii) \textit{Subspace/Sparsity-based Merging} (e.g., TIES \cite{yadav2023ties}, DARE \cite{yu2024language}): As mathematically rigorously proven in \textbf{Corollary \ref{cor:non_linear}}, these techniques directly apply the non-linear operator $\mathcal{M}(\cdot)$ to explicitly filter conflicting knowledge. This acts as an explicit regularizer that mechanically reduces the task heterogeneity from $\zeta_i^2$ to $\hat{\zeta}_i^2$. By explicitly optimizing the $\zeta_i^2$ parameter within our generalized $L_2$-stability framework, these methods strictly tighten both the stability and optimization error components. Recent methods like EMR-MERGING~\cite{huang2024ERM} and Twin-Merging~\cite{lu2024twin} follow the exact same mathematical mechanism of structural $\zeta_i^2$ reduction.
    
    \item \textbf{Post-calibration methods (e.g., Representation Surgery \cite{yang2024representation}):}  According to Theorem \ref{the:tight upper bound of excess error}, merging into a strictly shared parameter space inevitably leaves a residual optimization error ($\mathcal{E}_O$) driven by irreducible task heterogeneity ($\sum_{i=1}^N \lambda_i \zeta_i^2$). Post-calibration elegantly bypasses this bottleneck: by freezing the merged model $\boldsymbol{x}_{avg}$ to strictly cap the global stability penalty ($\mathcal{E}_G$), these methods introduce lightweight task-specific modules to absorb the residual $\mathcal{E}_O$. This safely minimizes the overall excess risk without triggering new algorithmic instability.
  \end{itemize}
\end{remark}

\begin{remark}\label{re:different_hyper}
  \textbf{(Quantitative Scaling Laws and Practical Fine-tuning Recommendations):}
  Our theoretical framework not only unifies merging algorithms but also provides \textbf{Quantitative Scaling Laws} governing the fine-tune-then-merge process. In Section \ref{sec:verification}, we provide a detailed mathematical analysis of how each hyperparameter (e.g., $\eta_l, b_i, K_i$) uniquely dictates the final generalization bound. Instead of heuristic recommendations, we offer exact formulas and theoretically validated guidelines for constructing merge-friendly experts in Appendix~\ref{app:hyperparameter_jieshi}. 
\end{remark}

\begin{remark}
\textbf{(Generalization to Adam.)}
    While Theorem \ref{eq:yizhi_upper bound of gradient norm} focuses on SGD for clarity, Theorem \ref{the:wang_etal} (Appendix) establishes a general framework. As discussed in \cite{wang2020tackling} and our Appendix (Eq.(\ref{eq:normal})), adaptive optimizers (Adam) can be modeled by modifying the effective step-size vector $\boldsymbol{a}_i$ (i.e., adaptive learning rates). This alters the coefficients in the bound but preserves the fundamental stability-optimization trade-off, ensuring the theory remains strictly applicable. Furthermore, experiments on ViT using Adam also corroborate the proposed theory.
\end{remark}

\section{Experiments}
\label{sec:experiments}

In this section, we empirically validate the theoretical insights from Section \ref{sec:theoretical_analysis}. Our goal is to demonstrate that our proposed stability-based framework is not merely explanatory, but acts as a predictive science. Specifically, we aim to verify whether the derived Quantitative Scaling Laws (Remark \ref{re:different_hyper}) accurately predict the behavior of the merged model $\boldsymbol{x}_{avg}$ under diverse finetuning hyperparameters. We analyze the effects of finetuning steps ($K_i$), batch size ($b_i$), learning rate ($\eta_l$), data ratio ($\alpha_i$), and the number of merged tasks ($N$). 
Our code is publicly available at \url{https://gitcode.com/tanganke/stability_model_merging}.

\subsection{Experimental Setup}
\label{sec:setup}

\begin{figure*}[t]
  \centering
  \includegraphics[width=\textwidth]{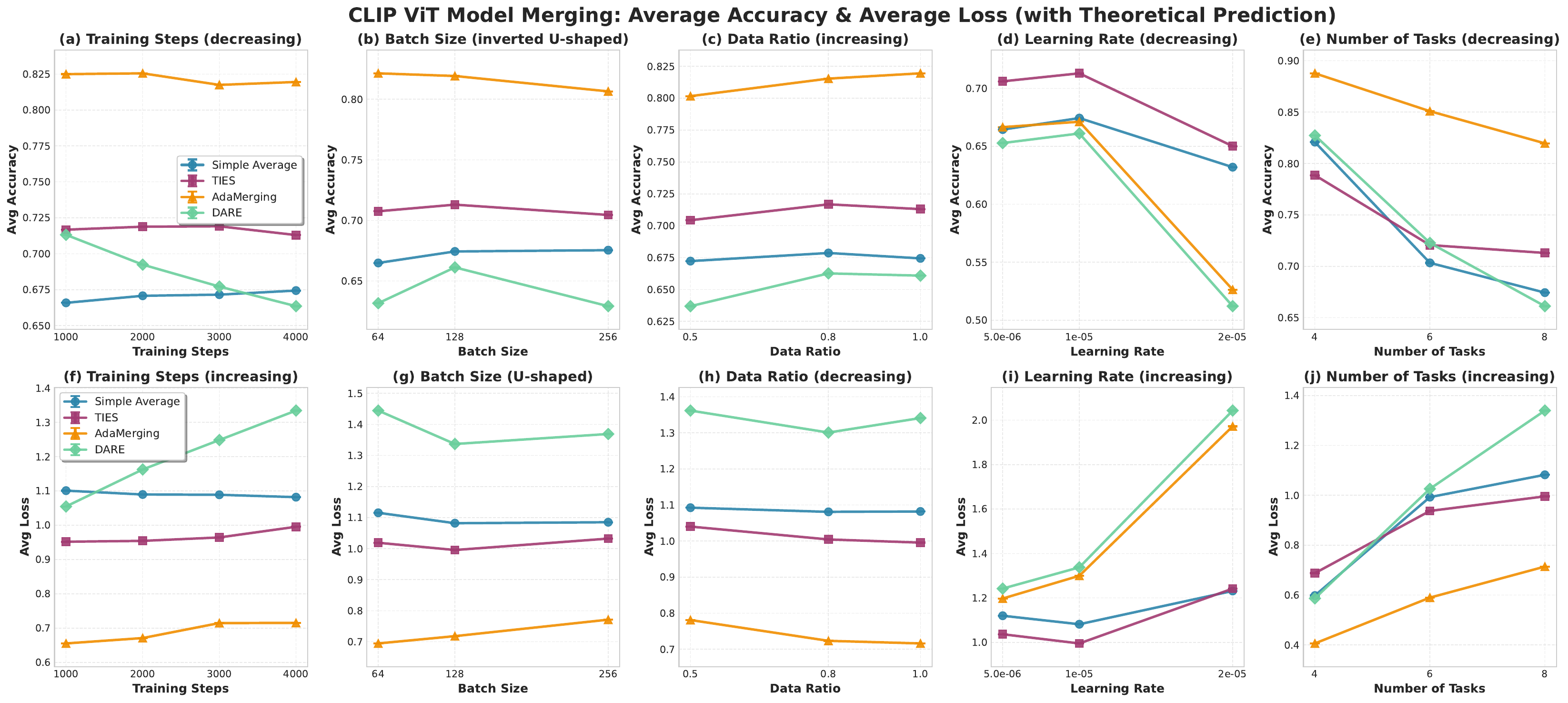}
  \caption{\textbf{Empirical Validation of Theoretical Predictions in Transformer (ViT).} The experiments evaluate four representative methods under different algorithmic hyperparameters in terms of average accuracy and average loss, both measured on $\boldsymbol{x}_{avg}$. Each subfigure also includes the theoretically predicted scaling trend derived from Theorem~\ref{the:tight upper bound of excess error}. As can be observed, linear and non-linear methods (TIES, DARE) closely follow the theoretical predictions, demonstrating the broad applicability of our framework to adaptive optimizers and modern architectures.}
  \label{fig:main_results_vit}
  \vspace{-1em}
\end{figure*}

\textbf{Datasets.}
Our experimental testbed consists of 20 diverse visual classification tasks, ranging from standard benchmarks like CIFAR-10 and SVHN to more complex, fine-grained challenges such as Stanford Cars and Food-101, as well as scene recognition with SUN397 and satellite imagery with EuroSAT. The complete dataset can be found in the Appendix \ref{app:exp_detail}. This variety ensures a thorough evaluation of model merging performance under varying degrees of task heterogeneity ($\zeta_i^2$). For each experiment, we evaluate each merged model on the joint test set—the union of all task test sets—providing a unified measure of its multi-task generalization.

\textbf{Models and Merging Methods.}
To rigorously test the universality of our bounds, we use three widely adopted backbones from the ResNet family—ResNet-18, ResNet-50, and ResNet-152—as well as the CLIP model based on the Vision Transformer (ViT) architecture. Each expert model is initialized from the same pretrained checkpoint, obtained from Hugging Face, and then finetuned on its respective task. During fine-tuning, we use SGD as the optimizer for ResNet-based models and Adam for ViT-based models. Unless otherwise specified as the variable under investigation, we use a standard set of hyperparameters for finetuning. For ResNet-based backbones, we adopt Simple Averaging, which is equivalent to Task Arithmetic \cite{TA2022editing} when the scaling factor is $1/N$. Crucially, for ViT-based frameworks, we evaluate more advanced non-linear merging methods, including TIES \cite{yadav2023ties}, DARE \cite{yu2024language}, and AdaMerging \cite{yang2023adamerging}. This design explicitly validates \textbf{Corollary \ref{cor:non_linear}} and our theoretical extensions to adaptive optimizers.

\textbf{Evaluation Protocol.}
To ensure robust and statistically significant results, we employ a randomized sampling procedure. For each experimental setting, we randomly sample 15 distinct groups of tasks from our pool of 20. We then perform the merging process for each group and report the \textit{average loss and accuracy} across these 15 trials. All experiments are conducted separately for each of the three ResNet backbones. For experiments based on the Transformer architecture, we evaluate on 8 tasks \cite{yang2023adamerging}. Unlike the ResNet setting, we fix the number of tasks to 8 in all experiments except those that explicitly study the effect of task count.

\begin{figure*}[t]
  \centering
  \includegraphics[width=\textwidth]{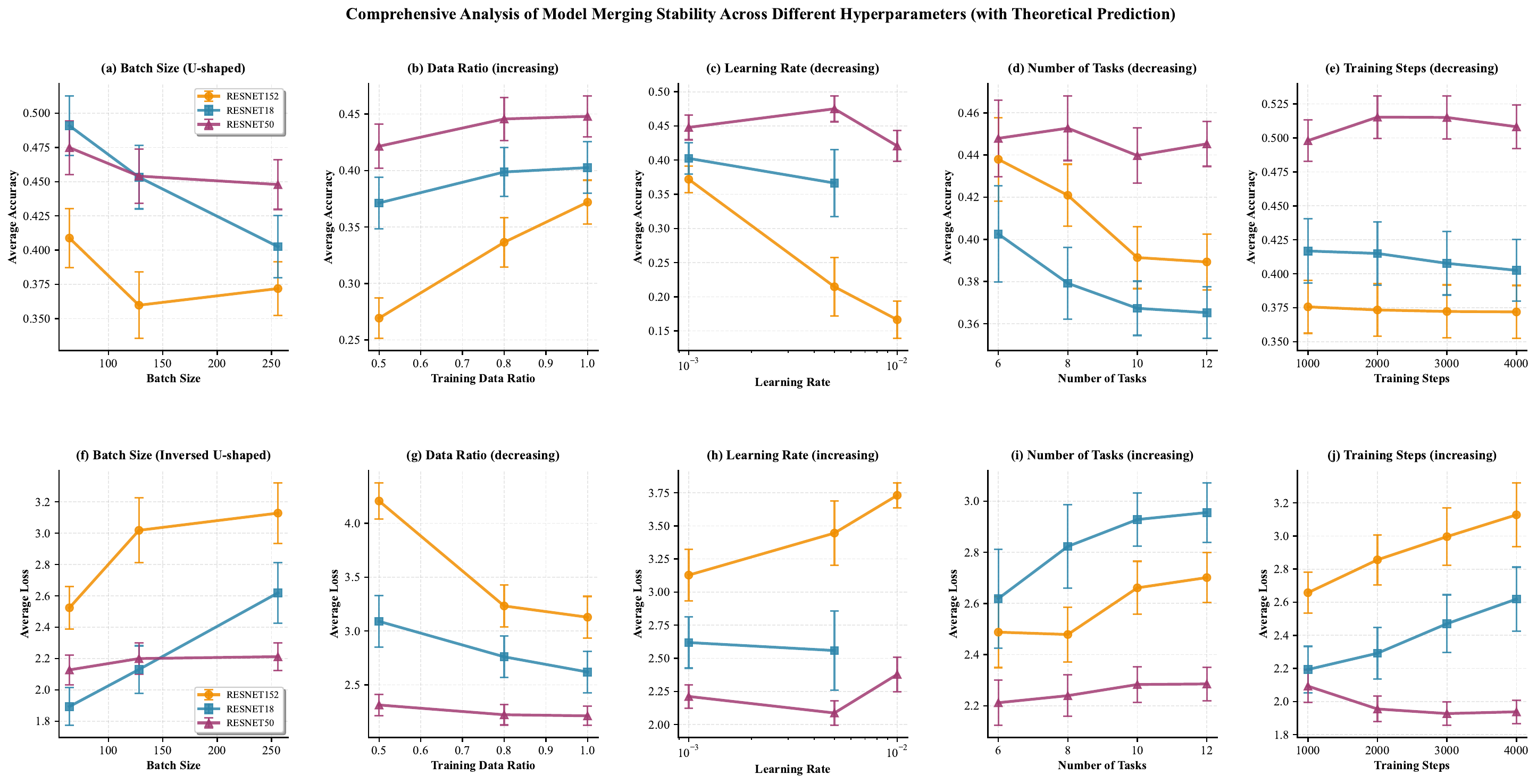}
  \caption{\textbf{Empirical Validation of Theoretical Predictions in ResNet.} We analyze the impact of key finetuning hyperparameters on the performance of the merged model ($\boldsymbol{x}_{avg}$). Error bars represent standard error over 15 independent random task group samplings. Each subplot corresponds to varying one hyperparameter while keeping others fixed. Performance is measured by accuracy (higher is better) and loss (lower is better) on the joint test set, averaged over 15 random task groups. The observed trends robustly align with the quantitative trade-offs predicted by our $L_2$-Stability-based generalization error bound in Theorem~\ref{the:tight upper bound of excess error}.}
  \label{fig:main_results}
  \vspace{-1em}
\end{figure*}

\subsection{Verifying Theoretical Predictions via Scaling Laws}
\label{sec:verification}
We now present the results of five targeted experiments, each designed to validate a specific scaling law mathematically derived from our theoretical framework. The results are visualized in Figure~\ref{fig:main_results_vit}\&\ref{fig:main_results}.

\subsubsection{Impact of Fine-tuning Steps (\texorpdfstring{$K_i$}{b-i})}\label{sec:exp_k_i}

\textbf{Quantitative Scaling Law (Optimization-Generalization Trade-off).}
As analyzed in Appendix~\ref{app:hyperparameter_jieshi}, the number of finetuning steps, $K_i$, presents a fundamental trade-off. Our bound in Theorem~\ref{the:tight upper bound of excess error} indicates that increasing $K_i$ reduces the optimization error component (specifically, the $\epsilon_{sgd}$ term, which decays with $\bar{K}$), but inflates the model stability component, which grows linearly with $K_i$. This suggests an optimal number of steps that balances under-fitting (large optimization error) and over-specialization (high instability). Consequently, the performance of the merged model depends on which of these competing factors dominates. We might observe performance first increasing and then decreasing with $K_i$, forming an inverted U-shape. Alternatively, if the stability cost grows faster than the optimization benefit from the start, performance could also decrease monotonically.

\textbf{Setup and Results.}
We merge models finetuned for $K_i \in \{1000, 2000, 3000, 4000\}$ steps, while holding other hyperparameters constant. As depicted in Figure~\ref{fig:main_results}(a), the empirical results show that the loss exhibits a monotonic increase or decrease, with accuracy changing inversely, which aligns with our theoretical predictions. Different models have distinct optimal values $K_i^*$. When the actual number of steps exceeds $K_i^*$, the loss rises because the increase in model instability dominates the overall excess error. Conversely, when the steps are fewer than $K_i^*$, the loss decreases as the reduction in optimization error becomes the dominant factor. These observations strongly support our theoretical prediction of the optimization–generalization trade-off governed by $K_i$.

\subsubsection{Impact of Batch Size \texorpdfstring{($b_i$)}{b-i}}\label{sec:b_i}

\textbf{Quantitative Scaling Law (Conditional Trade-off).}
Our theory highlights a mathematically dual role for the batch size $b_i$. A larger $b_i$ reduces stochastic gradient variance, accelerating the convergence of $\epsilon_{sgd}$. However, Theorem~\ref{the:bound_of_stablity} dictates that the stability penalty grows linearly with $b_i$ (scaled by $\frac{3b_i}{n_i}\zeta_i^2$), reflecting increased sensitivity to data perturbations under high task heterogeneity. Crucially, Theorem \ref{the:tight upper bound of excess error} reveals that this negative stability impact is structurally suppressed by the learning rate coefficient $\eta_l^2$. Thus, our scaling law predicts a \textit{conditional} behavior: in fine-tuning regimes with small learning rates, within an appropriate range, larger $b_i$ yields optimization benefits that overwhelmingly dominate the reduced stability cost.

\textbf{Setup and Results.}
We vary the batch size $b_i \in \{64, 128, 256\}$ for all finetuning processes. The results, depicted in Figure~\ref{fig:main_results_vit}\&\ref{fig:main_results}, show a consistent trend of improved performance as the batch size increases. As precisely formulated in our theoretical discussion, under our experimental setting with small learning rates ($\eta_l=0.001$ in ResNet and $\eta_l = 0.00001$ in ViT), the negative impact on model stability is heavily truncated by the $\eta_l^2$ term. Consequently, the reduction in optimization error ($\epsilon_{sgd}$) dictates the overall trend. This underscores the precision of our bound in capturing cross-hyperparameter dependencies.

\subsubsection{Impact of Learning Rate \texorpdfstring{($\eta_l$)}{eta}}\label{sec:eta}

\textbf{Quantitative Scaling Law (Quadratic Stability Penalty).}
The learning rate $\eta_l$ emerges in Theorem \ref{the:tight upper bound of excess error} as the most dominant factor controlling merging stability. The entire generalization penalty is scaled by $\eta_l^2$. Our framework provides a rigorous mathematical warning: while increasing $\eta_l$ accelerates local optimization, it triggers a \textit{quadratic explosion} ($\mathcal{O}(\eta_l^2)$) in the stability bound. Thus, the scaling law strictly forbids large learning rates, predicting that they will cause catastrophic generalization failure regardless of optimization convergence.

\textbf{Setup and Results.}
We finetune expert models using learning rates $\eta_l \in \{0.001, 0.005, 0.01\}$ in ResNet and $\{1e-5, 5e-6, 2e-5\}$ in ViT. The empirical results in Figure~\ref{fig:main_results_vit}\&\ref{fig:main_results} provide compelling evidence for this quadratic penalty. We observe a sharp, non-linear degradation in performance as the learning rate increases. Strikingly, when $\eta_l = 0.01$, all merged models fine-tuned on ResNet-18 completely collapsed. This confirms that maintaining a small learning rate is an absolute algorithmic prerequisite for preserving the $L_2$-stability required for successful model fusion.

\subsubsection{Impact of Data Ratio \texorpdfstring{($\alpha_i$)}{b-i}}\label{sec:data_ratio}

\textbf{Quantitative Scaling Law (Strict Monotonicity).}
The effect of the per-task dataset size, $n_i$ (controlled by data ratio $\alpha_i$), is unique in our equations. The dataset size $n_i$ appears exclusively in the denominator of the entire stability term, $\mathcal{O}(1/n_i)$. Unlike $K_i$ or $b_i$, varying $n_i$ introduces no competing effects in the optimization error. Consequently, our formula guarantees a "free lunch": increasing $n_i$ strictly and monotonically tightens the excess risk bound.

\textbf{Setup and Results.}
We simulate varying dataset sizes by using a fraction $\alpha_i \in \{0.5, 0.8, 1.0\}$ of the training data. As shown in Figure~\ref{fig:main_results_vit}\&\ref{fig:main_results}, the performance of the merged model improves monotonically as the data ratio increases. This result validates our mathematical formulation, confirming that scaling up training data unconditionally suppresses algorithmic sensitivity and yields inherently merge-friendly experts.

\subsubsection{Impact of Number of Tasks \texorpdfstring{($N$)}{}}\label{sec:num_of_task}

\textbf{Quantitative Scaling Law (Heterogeneity Accumulation).}
\looseness = -1
The number of merged tasks, $N$, introduces highly asymmetric competing effects in our bound. The optimization benefit improves extremely slowly at a rate of $\mathcal{O}(1/\sqrt{N})$, and its leading coefficient is typically small since finetuning begins from a highly capable pretrained checkpoint ($\boldsymbol{x}_0$). In stark contrast, the penalty terms associated with task heterogeneity ($\sum_{i=1}^N \lambda_i \zeta_i^2$) accumulate at least linearly with $N$. Our scaling law thus predicts that the rapidly growing heterogeneity penalty will invariably overwhelm the marginal optimization benefits, strictly degrading multi-task generalization as $N$ scales.

\textbf{Setup and Results.}
We vary the number of merged tasks $N \in \{6, 8, 10, 12\}$. In Figure~\ref{fig:main_results_vit}\&\ref{fig:main_results}(e), performance consistently degrades as more tasks are aggregated. This perfectly mirrors our analytical prediction regarding the accumulation of the variance term $\zeta_i^2$. This phenomenon also explicitly mathematically justifies the necessity of non-linear sparsification methods (e.g., TIES, DARE), which directly suppress $\zeta_i^2$ (as proven in \textbf{Corollary \ref{cor:non_linear}}) to successfully scale to a larger number of tasks.

\section{Conclusion}

In this paper, we established a pioneering $L_2$-stability-based generalization framework for model merging under heterogeneous hyperparameter environments. By rigorously decoupling the excess risk into optimization and generalization components, we successfully resolved the paradox inherent in applying traditional federated optimization theories to model fusion. Building upon this foundational framework, our contributions are twofold: 
(i) We theoretically unified both linear and state-of-the-art \textit{non-linear} merging algorithms (e.g., TIES, DARE), mathematically proving that sparsification and sign-consensus operators strictly tighten the excess error bound by explicitly suppressing task heterogeneity ($\zeta_i^2$).
(ii) We transitioned empirical heuristic guidelines into precise \textit{Quantitative Scaling Laws}, delineating exactly how finetuning hyperparameters dictate the fundamental optimization-generalization trade-off for the merged model $\boldsymbol{x}_{avg}$. 
Ultimately, these contributions bridge the critical gap between empirical crafts and predictive science in training merge-friendly expert models. In future work, we aim to mathematically relax key assumptions (e.g., incorporating non-smooth analysis for specific activation functions) to further generalize our bounds. Moreover, we anticipate that our mathematically derived scaling laws will inspire the next generation of tuning-free merging algorithms. For instance, directly employing our excess risk bound as a closed-form surrogate could guide the development of principled AutoML frameworks for automated hyperparameter selection in large-scale model merging.


\bibliographystyle{IEEEtran}
\bibliography{main.bib}

\begin{IEEEbiography}[{\includegraphics[width=1in,height=1.25in,clip,keepaspectratio]{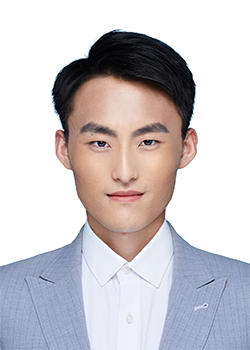}}]{Qinglun Li} received his Bachelor of Science in Mathematics from the School of Science at the University of Shanghai for Science and Technology. He is currently pursuing a doctoral degree at the National University of Defense Technology. His research interests include machine learning, decentralized federated learning, and non-convex optimization, with a particular focus on optimization and generalization theory for decentralized and federated systems. He has published multiple papers in leading journals and conferences, including IEEE Transactions on Pattern Analysis and Machine Intelligence, IEEE Transactions on Computers, NeurIPS, ICLR, CVPR, and Neural Networks. He has served as a reviewer for several top-tier conferences, including ICML, ICLR, NeurIPS, and CVPR.
\end{IEEEbiography}

\begin{IEEEbiography}
[{\includegraphics[width=1in,height=1.25in,clip,keepaspectratio]{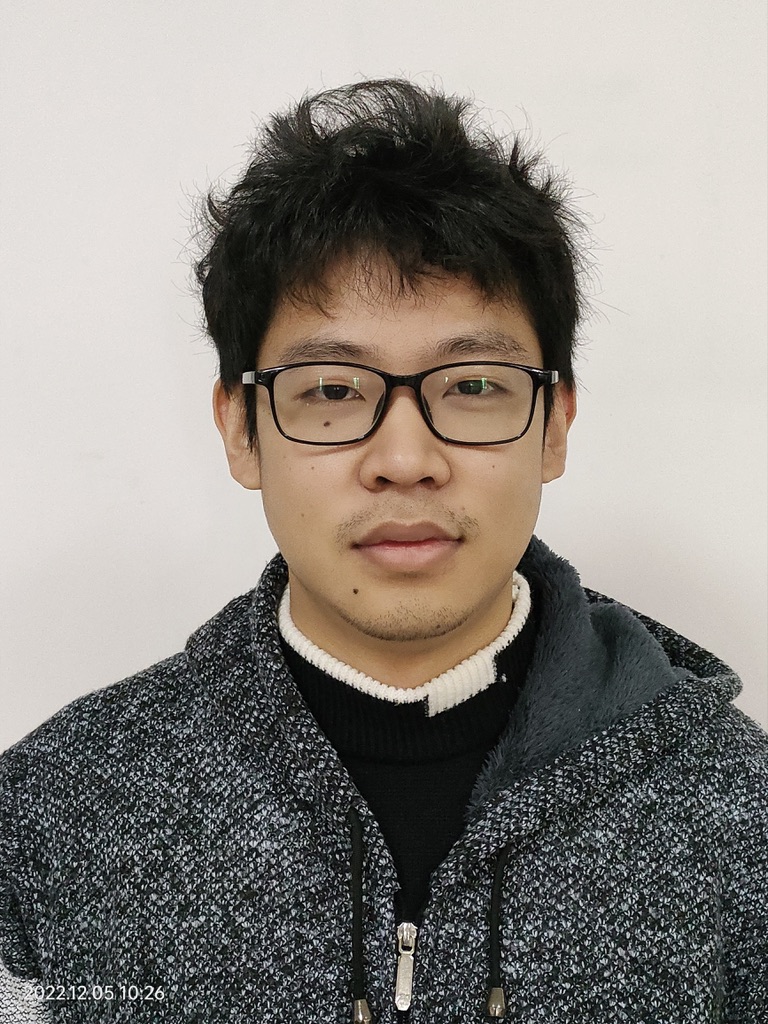}}]{Anke Tang} received his Ph.D. degree from the School of Computer Science, Wuhan University, under the supervision of Prof. Yong Luo and Li Shen. He received his Bachelor degree at the School of Physics and Technology, Wuhan University in 2020. His research interests include machine learning, transfer learning, continual learning and multi-task learning. He has published papers in proceedings at leading conferences, such as IJCAI, ICLR, ICML, and NeurIPS, and in top-tier journals including TPAMI and JMLR. He has also served as a reviewer for several top-tier conferences and journals.
\end{IEEEbiography}

\begin{IEEEbiography}[{\includegraphics[width=1in,height=1.25in, clip,keepaspectratio]{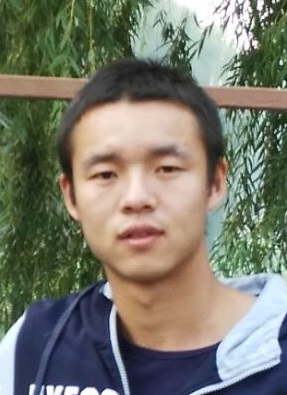}}]{Miao Zhang} is currently a research assistant at National University of Defense Technology, Changsha, China. He received his bachelor’s degree from the School of Information, University of Science and Technology of China in 2015, and received his master’s degree and Ph.D. from the College of Systems Engineering, National University of Defense Technology in 2017 and 2022, respectively. His research interests include intelligence optimization, resource management, private computing, federated learning and reinforcement learning.
\end{IEEEbiography}

\begin{IEEEbiography}[{\includegraphics[width=1in,height=1.25in,clip,keepaspectratio]{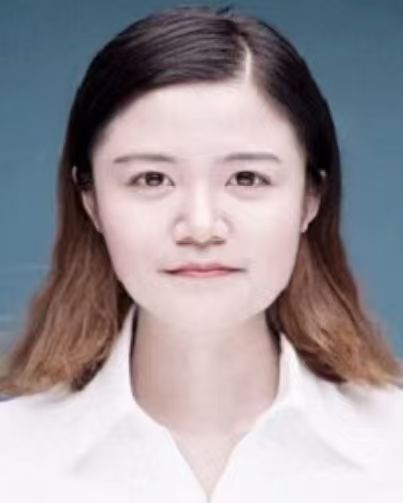}}]{Mengzhu Wang}
received the master's degree from Chongqing University, Chongqing, China, in 2018. She is currently pursuing the Ph.D. degree with the Science and Technology on Parallel and Distributed Laboratory, School of Computer Science, National University of Defense Technology, Changsha, China. She is currently an Associate Professor with the Hebei University of Technology. Her current research interests include transfer learning and computer vision.
\end{IEEEbiography}

\begin{IEEEbiography}[{\includegraphics[width=1in,height=1.25in, clip,keepaspectratio]{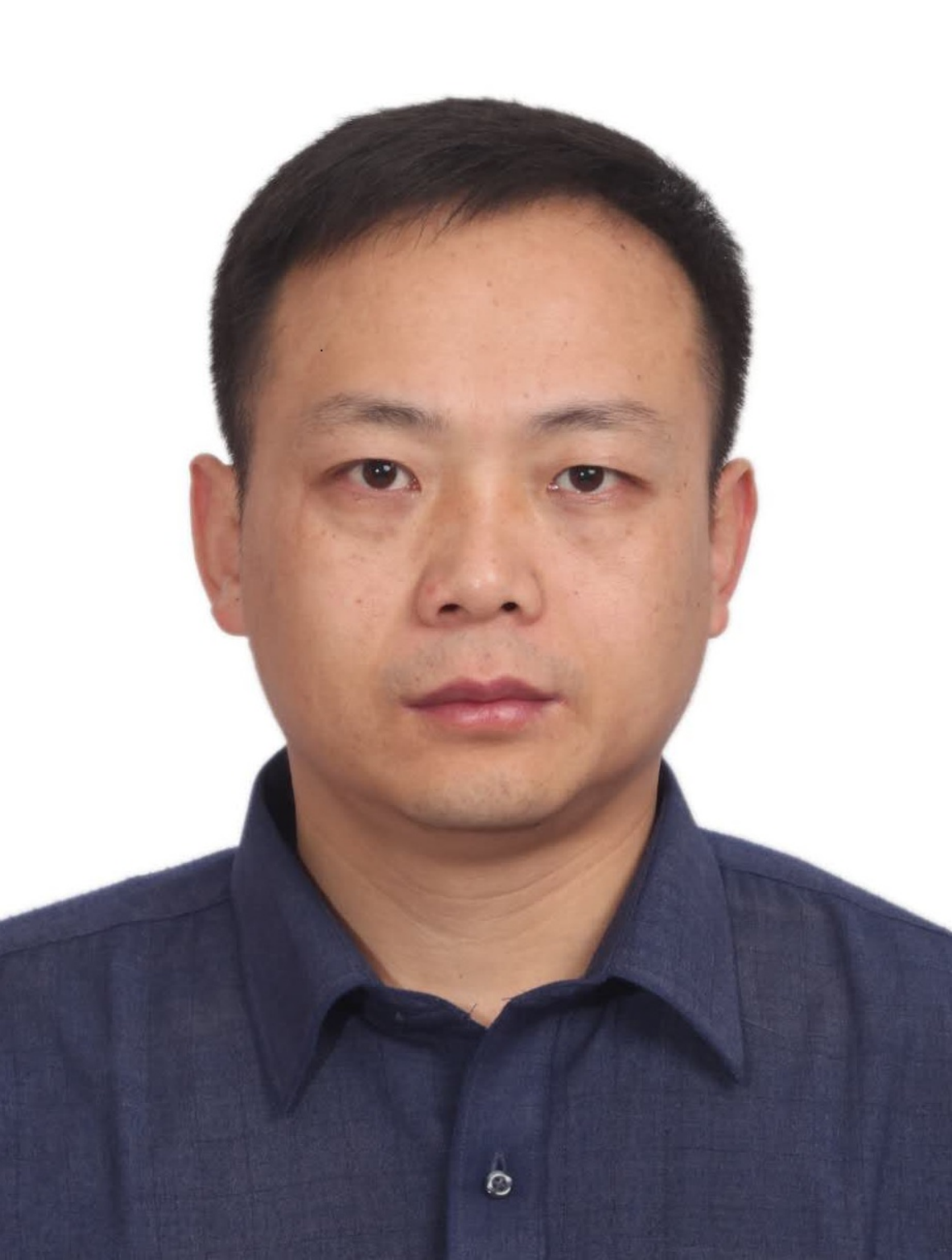}}]{Quanjun yin} is currently a professor with the College of Systems Engineering, National University of Defense Technology, China. His research interests include cognitive process modelling, qualitative spatial reasoning and planning, cooperation and negotiation, cloud-based simulation, and edge computing.
\end{IEEEbiography}

\begin{IEEEbiography}[{\includegraphics[width=1in,height=1.25in,clip,keepaspectratio]{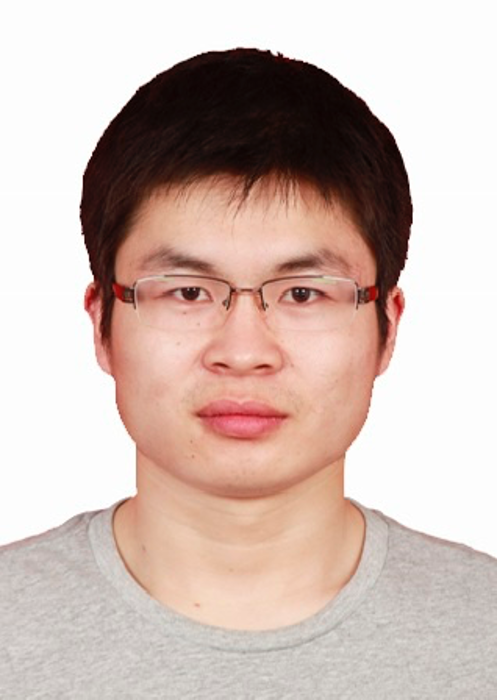}}]{Li Shen} is currently an associate professor at Sun Yat-sen University. Previously, he was a research scientist at JD Explore Academy, Beijing, and a senior researcher at Tencent AI Lab, Shenzhen. He received his bachelor's degree and Ph.D. from the School of Mathematics, South China University of Technology. His research interests include efficient deep learning, efficient reinforcement learning, optimization and deep learning theory. He has served as the senior program committee for AAAI and area chairs for ICML, NeurIPS, ICLR, CVPR, and ACMMM. He is also the associate editor for IEEE Transactions on Pattern Analysis and Machine Intelligence, IEEE Transactions on Knowledge and Data Engineering, and  IEEE Transactions on Multimedia.
\end{IEEEbiography}


\onecolumn

\appendices


\section{Appendix}\label{app:rheory}
In this part, we provide the supplementary materials to prove the main theorem.

\subsection{Preliminary Lemmas}

Then, given a learning algorithm $\mathcal{A}(\cdot)$ and a training dataset $\mathcal{D}$, we denote $\boldsymbol{x} = \mathcal{A}(\mathcal{D})$  as the model generated by the method $\mathcal{A}$ with given data. Then, the upper bound of the model's generalization gap is established in the following theorem:

\begin{lemma}[\textbf{Generalization via on-average model stability}] \label{lemma:ob-avg-gen}
  Let $\mathcal{D}, \tilde{\mathcal{D}}, \mathcal{D}^{(i)}$ be constructed as Definition \ref{def:perturbed dataset}. Let $\gamma > 0$. If for any $z$, the function $f(\boldsymbol{x}; z)$ is nonnegative and $L$-smooth, then
  \begin{equation*}
    \mathbb{E}_{\mathcal{D}, \mathcal{A}}[F(\mathcal{A}(\mathcal{D})) - f(\mathcal{A}(\mathcal{D}))] \leq \frac{1}{2\gamma} \mathbb{E}_{\mathcal{D}, \mathcal{A}}[\|\nabla f(\mathcal{A}(\mathcal{D}))\|^2] + \frac{L + \gamma}{2} \frac{1}{N} \sum_{i=1}^N \mathbb{E}_{\mathcal{D}, \tilde{\mathcal{D}}, \mathcal{A}}[\|\mathcal{A}(\mathcal{D}^{(i)}) - \mathcal{A}(\mathcal{D})\|^2].
  \end{equation*}

  \begin{proof}
    According to the definition of generalization error, we have:
    \begin{equation*}
      \begin{aligned}
         & \mathbb{E}_{\mathcal{D,A}}[F(\mathcal{A}(\mathcal{D}))-f(\mathcal{A}(\mathcal{D}))]                                                                                                                                                                                                                                   \\
         & =\mathbb{E}_{\mathcal{D},\mathcal{A}}\left[\frac{1}{N}\sum_{i=1}^{N}\left(F(\mathcal{A}(\mathcal{D}))\right)\right]-\mathbb{E}_{\mathcal{D},\mathcal{A}}\left[f(\mathcal{A}(\mathcal{D}))\right]                                                                                                                      \\
         & =\mathbb{E}_{\mathcal{D},\mathcal{A}}\left[\frac{1}{N}\sum_{i=1}^{N}\mathbb{E}_{\tilde{z}_{i}}\left[\ell(\mathcal{A}(\mathcal{D}),\tilde{z}_{i})\right]\right]-\mathbb{E}_{\mathcal{D},\mathcal{A}}\left[f(\mathcal{A}(\mathcal{D}))\right]                                                                           \\
         & =\mathbb{E}_{\mathcal{D},\mathcal{A}}\left[\frac{1}{N}\sum_{i=1}^{N}\mathbb{E}_{\tilde{z}_{i}}\left[\ell(\mathcal{A}(\mathcal{D}),\tilde{z}_{i})\right]\right]-\mathbb{E}_{\mathcal{D},\mathcal{A}}\left[\frac{1}{N}\sum_{i=1}^{N}\ell(\mathcal{A}(\mathcal{D});z_{i})\right]                                         \\
         & =\mathbb{E}_{\mathcal{D},\mathcal{A}}\left[\frac{1}{N}\sum_{i=1}^{N}\mathbb{E}_{\tilde{\mathcal{D}}}\left[\ell(\mathcal{A}(\mathcal{D}),\tilde{z}_{i})\right]\right]-\frac{1}{N}\sum_{i=1}^{N}\mathbb{E}_{\mathcal{D},\tilde{\mathcal{D}},\mathcal{A}}\left[\ell(\mathcal{A}(\mathcal{D}^{(i)});\tilde{z}_{i})\right] \\
         & =\mathbb{E}_{\mathcal{D},\tilde{\mathcal{D}},\mathcal{A}}\left[\frac{1}{N}\sum_{i=1}^{N}\left(\ell(\mathcal{A}(\mathcal{D}),\tilde{z}_{i})-\ell(\mathcal{A}(\mathcal{D}^{(i)});\tilde{z}_{i})\right)\right].
      \end{aligned}
    \end{equation*}
    Since $\ell(x; z)$ is $L$-smooth, i.e., $\forall \boldsymbol{x}, \boldsymbol{y} \in \mathcal{X}, \forall z \in \mathcal{Z}, \|\nabla \ell(\boldsymbol{x}; z) - \nabla \ell(\boldsymbol{y}; z)\| \leq L \|\boldsymbol{x} - \boldsymbol{y}\|$, and by the definition of empirical risk $f(\boldsymbol{x}) = \frac{1}{N} \sum_{i=1}^N \ell(\boldsymbol{x}; z_j)$, we have
    \begin{align*}
      \|\nabla f(\boldsymbol{x}) - \nabla f(\boldsymbol{y})\| & = \left\|\frac{1}{N} \sum_{i=1}^N \nabla \ell(\boldsymbol{x}; z_j) - \frac{1}{N} \sum_{i=1}^N \nabla \ell(\boldsymbol{y}; z_j) \right\| \\
                                                              & \leq \frac{1}{N} \sum_{i=1}^N \|\nabla \ell(\boldsymbol{x}; z_j) - \nabla \ell(\boldsymbol{y}; z_j)\|                                   \\
                                                              & \leq L \|\boldsymbol{x} - \boldsymbol{y} \|
    \end{align*}

    Therefore, the empirical risk is also $L$-smooth, and we have the following inequality.

    \begin{align*}
      \mathbb{E}_{\mathcal{D},\mathcal{A}}[F(\mathcal{A}(\mathcal{D}))-f(\mathcal{A}(\mathcal{D}))] & \leq\mathbb{E}_{\mathcal{D},\tilde{\mathcal{D}},\mathcal{A}}\left[\langle\mathcal{A}(\mathcal{D}^{(i)})-\mathcal{A}(\mathcal{D}),\frac{1}{N}\sum_{i=1}^N\nabla\ell(\mathcal{A}(\mathcal{D});\tilde{z}_j)\rangle \right] \\
                                                                                                    & \quad + \frac{L}{2}\mathbb{E}_{\mathcal{D},\tilde{\mathcal{D}},\mathcal{A}}\left[\left\|\mathcal{A}(\mathcal{D}^{(i)})-\mathcal{A}(\mathcal{D})\right\|^2\right]
    \end{align*}

    Using Cauchy’s inequality, for any $\gamma > 0$, we have
    \begin{align*}
       & \mathbb{E}_{\mathcal{D},\tilde{\mathcal{D}},\mathcal{A}}\left[\langle\mathcal{A}(\mathcal{D}^{(i)})-\mathcal{A}(\mathcal{D}),\frac{1}{N}\sum_{i=1}^{N}\nabla\ell(\mathcal{A}(\mathcal{D});\tilde{z}_{i})\rangle\right]  \leq\mathbb{E}_{\mathcal{D},\tilde{\mathcal{D}},\mathcal{A}}\|\mathcal{A}(\mathcal{D}^{(i)})-\mathcal{A}(\mathcal{D})\|\cdot\|\nabla f(\mathcal{A}(\mathcal{D}))\| \\
       & \hspace{15em} \leq\frac{\gamma}{2}\mathbb{E}_{\mathcal{D},\tilde{\mathcal{D}},\mathcal{A}}\|\mathcal{A}(\mathcal{D}^{(i)})-\mathcal{A}(\mathcal{D})\|^{2}+\frac{1}{2\gamma}\mathbb{E}_{\mathcal{D},\tilde{\mathcal{D}},\mathcal{A}}\|\nabla f(\mathcal{A}(\mathcal{D}))\|^{2}
    \end{align*}
    By further simplification, we obtain the final result.
    \begin{align*}
      \mathbb{E}_{\mathcal{D},\mathcal{A}}[F(\mathcal{A}(\mathcal{D}))-f(\mathcal{A}(\mathcal{D}))]\leq\frac{L+\gamma}{2}\underbrace{\frac{1}{N}\sum_{i=1}^N\mathbb{E}_{\mathcal{D},\tilde{\mathcal{D}},\mathcal{A}}[\|\mathcal{A}(\mathcal{D}^{(i)})-\mathcal{A}(\mathcal{D})\|^2]}_{\text{On-average Model stability}}+\frac{1}{2\gamma}\underbrace{\mathbb{E}_{\mathcal{D},\mathcal{A}}\|\nabla f(\mathcal{A}(\mathcal{D}))\|^2}_{\text{Gradient norm}}
    \end{align*}

  \end{proof}

\end{lemma}

In fact, we modified the proof of the lemma from \cite{lei2020fine}, replacing the $f(A(\mathcal{D}))$ on the right-hand side with a gradient $\mathbb{E}_{A,\mathcal{D}}[\|\nabla f(A(\mathcal{D}))\|^2]$. This adjustment better captures the impact of data heterogeneity on the generalization error.
With this lemma, obtaining the desired generalization bound reduces to controlling the $l_2$ on-average model stability of the decentralized algorithm $A$.

\textbf{Step 1: Proof of Local Model Stability:}
\begin{lemma}\label{lemma2:local stability}
  Letting non-negative objectives $f_i, \forall i \in [N]$ satisfy L-smoothness and Assumptions \ref{ass:bound SG}, \ref{ass:bound_hetero}. We suppose the $i$-th task preserves dataset $\mathcal{D}_i$ and $|\mathcal{D}_i| = n_i$ samples. Its perturbed dataset $\mathcal{D}_i^{(j)}$ has the perturbed sample $\tilde{z}_j \in \mathcal{D}_i$ with probability $1$. If the $i$-th task conducts $K_i$ mini-batch SGD steps with batch-size $b_i$, and non-increasing learning rate $\eta_t = \Theta\left(\frac{1}{KL}\right)$, we prove the stability of local iteration on the $i$-th task as
  \begin{equation}\label{eq:local stability}
    \mathbb{E}_{\mathcal{D}_i,\mathcal{D}_i^{\prime}}\|\boldsymbol{x}_i^{K_i}-\tilde{\boldsymbol{x}}_i^{K_i}\|^2 \leq 16K\eta_l^2\left(\frac{\sigma_i^2}{n_i}+\frac{3b_i\zeta_i^2}{n_i}\right)
  \end{equation}

  \begin{proof}
    Noting that the dataset $\mathcal{D}_i$ and the perturbed dataset $\tilde{\mathcal{D}}_i$ differ by at most one sample, there are two possible cases to consider when running local mini-batch size SGD.

    \textbf{First Case:} In the first case, the local mini-batch size SGD does not select the perturbed sample from either $\mathcal{D}_i$ or $\tilde{\mathcal{D}}_i$, in which case we have

    \begin{align*}
       & \mathbb{E}[\|\boldsymbol{x}_{i}^{k+1}-\tilde{\boldsymbol{x}}_{i}^{k+1}\|^{2}|\tilde{z}\notin\xi]                         \\
       & \leq\mathbb{E}\|\boldsymbol{x}_{i}^{k}-\tilde{\boldsymbol{x}}_{i}^{k}-\eta_{l}(\boldsymbol{g}_{i}^{k}-\tilde{\boldsymbol{g}}_{i}^{k})\|^{2}     \\
       & \leq\mathbb{E}\left[\|\boldsymbol{x}_{i}^{k}-\tilde{\boldsymbol{x}}_{i}^{k}\|^{2}+\eta_{l}^{2}\|\boldsymbol{g}_{i}^{k}-\tilde{\boldsymbol{g}}_{i}^{k}\|^{2}-2\eta_{l}\langle\boldsymbol{x}_{i}^{k}-\tilde{\boldsymbol{x}}_{i}^{k},\boldsymbol{g}_{i}^{k}-\tilde{\boldsymbol{g}}_{i}^{k}\rangle\right]                            \\
       & \leq\mathbb{E}\left[\|\boldsymbol{x}{i}^{k}-\tilde{\boldsymbol{x}}_{i}^{k}\|^{2}+\eta_{l}^{2}\|\boldsymbol{g}_{i}^{k}-\tilde{\boldsymbol{g}}_{i}^{k}\|^{2}-2\eta_{l}\langle\boldsymbol{x}_{i}^{k}-\tilde{\boldsymbol{x}}_{i}^{k},\nabla f_{i}(\boldsymbol{x}_{i}^{k})-\nabla f_{i}(\tilde{\boldsymbol{x}}_{i}^{k})\rangle\right] \\
       & \leq\mathbb{E}\left[\|\boldsymbol{x}_{i}^{k}-\tilde{\boldsymbol{x}}_{i}^{k}\|^{2}+\eta_{l}^{2}\|\boldsymbol{g}_{i}^{k}-\tilde{\boldsymbol{g}}_{i}^{k}\|^{2}+2\eta_{l}L\|\boldsymbol{x}_{i}^{k}-\tilde{\boldsymbol{x}}_{i}^{k}\|^{2}\right]\\
       & \leq(1+2\eta_{l}L)\mathbb{E}\|\boldsymbol{x}_{i}^{k}-\tilde{\boldsymbol{x}}_{i}^{k}\|^{2}+\eta_{l}^{2}\mathbb{E}\|\boldsymbol{g}_{i}^{k}-\tilde{\boldsymbol{g}}_{i}^{k}\|^{2}     \\
       & =(1+2\eta_{l}L)\mathbb{E}\|\boldsymbol{x}_{i}^{k}-\tilde{\boldsymbol{x}}_{i}^{k}\|^{2}+\eta_{l}^{2}\mathbb{E}\|\boldsymbol{g}_{i}^{k}\pm\nabla f_{i}(\boldsymbol{x}_{i}^{k})-\tilde{\boldsymbol{g}}_{i}^{k}\pm\nabla f_{i}(\tilde{\boldsymbol{x}}_{i}^{k})\|^{2}   \\
       & \overset{(i)}{\operatorname*{\leq}}(1+2\eta_lL)\mathbb{E}\|\boldsymbol{x}_i^{k}-\tilde{\boldsymbol{x}}_i^{k}\|^2+2\eta_l^2\mathbb{E}\|\boldsymbol{g}_i^{k}-\nabla f_i(\boldsymbol{x}_i^{k})-\tilde{\boldsymbol{g}}_i^{k}+\nabla f_i(\tilde{\boldsymbol{x}}_i^{k})\|^2     \\
       & +2\eta_{l}^{2}\mathbb{E}\|\nabla f_{i}(\boldsymbol{x}_{i}^{k})-\nabla f_{i}(\tilde{\boldsymbol{x}}_{i}^{k})\|^{2}  \\
       & \leq (1+2\eta_lL)\mathbb{E}\|\boldsymbol{x}_i^{k}-\tilde{\boldsymbol{x}}_i^{k}\|^2+2\eta_l^2\mathbb{E}\|\boldsymbol{g}_i^{k}-\nabla f_i(\boldsymbol{x}_i^{k})\|^2+2\eta_l^2\mathbb{E}\|\tilde{\boldsymbol{g}}_i^{k}-\nabla f_i(\tilde{\boldsymbol{x}}_i^{k})\|^2  \\
       & +2\eta_l^2\mathbb{E}\|\nabla f_i(\boldsymbol{x}_i^{k})-\nabla f_i(\tilde{\boldsymbol{x}}_i^{k})\|^2     \\& \leq(1+2\eta_{l}L+2\eta_{l}^{2}L^{2})\mathbb{E}\|\boldsymbol{x}_{i}^{k}-\tilde{\boldsymbol{x}}_{i}^{k}\|^{2}+\frac{4\eta_{l}^{2}\sigma_{i}^{2}}{b_i}
    \end{align*}

    where we note that the stochastic gradients $\boldsymbol{g}_i^{k}$ and $\widetilde{\boldsymbol{g}}_i^{k}$ only differ in model parameters ($\boldsymbol{x}_i^k$ and $\tilde{\boldsymbol{x}}_i^k$) and the mini batch data samples are identical in $(i)$.

    \textbf{Second Case:} In the second case, local mini-batch SGD samples a batch of data that involves the perturbed sample $\tilde{z}$ from $\mathcal{D}_i$ and $\mathcal{D}_i'$, which happens with probability $b_i/n_i$. Analogously, we have
    \begin{equation}
      \begin{aligned}
         & \mathbb{E}[\|\boldsymbol{x}_{i}^{k+1}-\tilde{\boldsymbol{x}}_{i}^{k+1}\|^{2}|\tilde{z}\in\xi]                                                                                                                                                                                                                                            \\
         & \leq\mathbb{E}\|\boldsymbol{x}_{i}^{k}-\tilde{\boldsymbol{x}}_{i}^{k}-\eta_{l}(\boldsymbol{g}_{i}^{k}-\tilde{\boldsymbol{g}}_{i}^{k})\|^{2}                                                                                                                                                                                              \\
         & \leq\mathbb{E}\left[\|\boldsymbol{x}_{i}^{k}-\tilde{\boldsymbol{x}}_{i}^{k}\|^{2}+\eta_{l}^{2}\|\boldsymbol{g}_{i}^{k}-\tilde{\boldsymbol{g}}_{i}^{k}\|^{2}-2\eta_{l}\langle\boldsymbol{x}_{i}^{k}-\tilde{\boldsymbol{x}}_{i}^{k},\boldsymbol{g}_{i}^{k}-\tilde{\boldsymbol{g}}_{i}^{k}\rangle\right]                                    \\
         & \leq\mathbb{E}\left[\|\boldsymbol{x}_{i}^{k}-\tilde{\boldsymbol{x}}_{i}^{k}\|^{2}+\eta_{l}^{2}\|\boldsymbol{g}_{i}^{k}-\tilde{\boldsymbol{g}}_{i}^{k}\|^{2}-2\eta_{l}\langle\boldsymbol{x}_{i}^{k}-\tilde{\boldsymbol{x}}_{i}^{k},\nabla f_{i}(\boldsymbol{x}_{i}^{k})-\nabla\tilde{f}_{i}(\tilde{\boldsymbol{x}}_{i}^{k})\rangle\right] \\
         & \leq\mathbb{E}\bigg[\|\boldsymbol{x}_{i}^{k}-\tilde{\boldsymbol{x}}_{i}^{k}\|^{2}+\eta_{l}^{2}\|\boldsymbol{g}_{i}^{k}-\tilde{\boldsymbol{g}}_{i}^{k}\|^{2}                                                                                                                                                                              \\
         & -2\eta\langle\boldsymbol{x}_{i}^{k}-\widetilde{\boldsymbol{x}}_{i}^{k},\frac{1}{n_{i}}\sum_{j=1}^{n_{i}}\left(\nabla\ell(\boldsymbol{x}_{i}^{k};z_{j})-\nabla\ell(\tilde{\boldsymbol{x}}_{i}^{k};z_{j})\right)+\nabla\ell(\tilde{\boldsymbol{x}}_{i}^{k};z)-\nabla\ell(\tilde{\boldsymbol{x}}_{i}^{k};\tilde{z}))\rangle\bigg]           \\
         & \overset{(i)}{\leq}\mathbb{E}\left[\|\boldsymbol{x}_i^{k}-\tilde{\boldsymbol{x}}_i^{k}\|^2+\eta_l^2\|g_i^{k}-\tilde{g}_i^{k}\|^2-2\eta_l\langle \boldsymbol{x}_i^{k}-\tilde{\boldsymbol{x}}_i^{k},\nabla f_i(\boldsymbol{x}_i^{k})-\nabla f_i(\tilde{\boldsymbol{x}}_i^{k})\rangle\right]                                                \\
         & \leq\mathbb{E}\left[\|\boldsymbol{x}_{i}^{k}-\tilde{\boldsymbol{x}}_{i}^{k}\|^{2}+\eta_{l}^{2}\|\boldsymbol{g}_{i}^{k}-\tilde{\boldsymbol{g}}_{i}^{k}\|^{2}+2\eta_{l}L\|\boldsymbol{x}_{i}^{k}-\tilde{\boldsymbol{x}}_{i}^{k}\|^{2}\right]                                                                                               \\
         & \leq(1+2\eta_{l}L)\mathbb{E}\|\boldsymbol{x}_{i}^{k}-\tilde{\boldsymbol{x}}_{i}^{k}\|^{2}+\eta_{l}^{2}\mathbb{E}\|\boldsymbol{g}_{i}^{k}-\tilde{\boldsymbol{g}}_{i}^{k}\|^{2}                                                                                                                                                            \\
         & \overset{(ii)}{\leq}(1+2\eta_lL)\mathbb{E}\|\boldsymbol{x}_i^{k}-\tilde{\boldsymbol{x}}_i^{k}\|^2+\eta_l^2\mathbb{E}\|\boldsymbol{g}_i^{k}\pm\nabla f_i(\boldsymbol{x}_i^{k})-\tilde{\boldsymbol{g}}_i^{k}\pm\nabla\tilde{f}_i(\tilde{\boldsymbol{x}}_i^{k})\|^2                                                                         \\
         & \leq(1+2\eta_{l}L)\mathbb{E}\|\boldsymbol{x}_{i}^{k}-\tilde{\boldsymbol{x}}_{i}^{k}\|^{2}+\frac{4\eta_{l}^{2}\sigma_{i}^{2}}{b_i}+2\eta_{l}^{2}\mathbb{E}\|\nabla f_{i}(\boldsymbol{x}_{i}^{k})-\nabla\tilde{f}_{i}(\tilde{\boldsymbol{x}}_{i}^{k})\|^{2}
      \end{aligned}
    \end{equation}
    where $(i)$ utilizes the fact that $z, \tilde{z} \stackrel{i.i.d.}{\sim} \mathcal{P}_i$, implying that
    $\mathbb{E}_{z, \tilde{z}} \left[ \nabla \ell(\tilde{\boldsymbol{x}}_i^{k}; z) - \nabla \ell(\tilde{\boldsymbol{x}}_i^{k}; \tilde{z}) \right] = 0,$
    due to the expectation being taken over the local data distribution. Since the perturbed samples $z$ and $\tilde{z}$ are independently and identically distributed (i.i.d.), and under Assumption~\ref{ass:smooth}, which characterizes the smoothness of the local data distribution, the above holds.
    Moreover, Assumption \ref{ass:bound SG} is assumed to hold for SGD with batch size 1, i.e.,
    $\|\nabla \ell(\boldsymbol{x}; z) - \nabla f_i(\boldsymbol{x})\| \leq \sigma_i^2, \quad \forall \boldsymbol{x} \in \mathcal{X},\, z \in \mathcal{D}_i \sim \mathcal{P}_i.$
    We define the virtual local objective as
    $\tilde{f}_i(\tilde{\boldsymbol{x}}_i^{k}) := \frac{1}{|\mathcal{D}_i'|} \sum_{z \in \mathcal{D}_i'} \ell(\tilde{\boldsymbol{x}}_i^{k}; z),$
    which differs from $f_i(\tilde{\boldsymbol{x}}_i^{k}) = \frac{1}{|\mathcal{D}_i|} \sum_{z \in \mathcal{D}_i} \ell(\tilde{\boldsymbol{x}}_i^{k}; z)$ by considering a perturbed sample.
    In step $(ii)$, we use the fact that both $\tilde{f}_i(\tilde{\boldsymbol{x}}_i^{k})$ and the corresponding stochastic gradient $\tilde{\boldsymbol{g}}_i^{k}$ satisfy Assumption \ref{ass:bound SG}. In addition, $\tilde{f}_i(\boldsymbol{x})$ also satisfies Assumption \ref{ass:bound_hetero} regarding its alignment with the global objective $f$, in the same way as $f_i(\boldsymbol{x})$ does, for all $\boldsymbol{x} \in \mathcal{X}$. This holds since $z$ and $\tilde{z}$ are drawn i.i.d. from $\mathcal{P}_i$, for every $i \in [N]$. Based on the above discussion, we conclude that
    \begin{align*}
       & \mathbb{E}[\|\boldsymbol{x}_{i}^{k+1}-\tilde{\boldsymbol{x}}_{i}^{k+1}\|^{2}|\tilde{z}\in\xi]                                                                                                                                                                                                                                       \\
       & \leq(1+2\eta_{l}L)\mathbb{E}\|\boldsymbol{x}_{i}^{k}-\tilde{\boldsymbol{x}}_{i}^{k}\|^{2}+\frac{4\eta_{l}^{2}\sigma_{i}^{2}}{b_i}+2\eta_{l}^{2}\mathbb{E}\|\nabla f_{i}(\boldsymbol{x}_{i}^{k})-\nabla\tilde{f}_{i}(\tilde{\boldsymbol{x}}_{i}^{k})\|^{2}                                                                           \\
       & \leq(1+2\eta_{l}L)\mathbb{E}\|\boldsymbol{x}_{i}^{k}-\tilde{\boldsymbol{x}}_{i}^{k}\|^{2}+\frac{4\eta_{l}^{2}\sigma_{i}^{2}}{b_i}+2\eta_{l}^{2}\mathbb{E}\|\nabla f_{i}(\boldsymbol{x}_{i}^{k}) \pm \nabla f(\boldsymbol{x}_i^k)-\nabla\tilde{f}_{i}(\tilde{\boldsymbol{x}}_{i}^{k}) \pm \nabla f(\tilde{\boldsymbol{x}}_i^k)\|^{2} \\
       & \leq(1+2\eta_{l}L)\mathbb{E}\|\boldsymbol{x}_{i}^{k}-\tilde{\boldsymbol{x}}_{i}^{k}\|^{2}+\frac{4\eta_{l}^{2}\sigma_{i}^{2}}{b_i}                                                                                                                                                                                                   \\
       & \quad + 2\eta_{l}^{2}\mathbb{E}\|\nabla f_{i}(\boldsymbol{x}_{i}^{k}) - \nabla f(\boldsymbol{x}_i^k)  + \nabla f(\tilde{\boldsymbol{x}}_i^k)- \nabla\tilde{f}_{i}(\tilde{\boldsymbol{x}}_{i}^{k}) + \nabla f(\boldsymbol{x}_i^k) - \nabla f(\tilde{\boldsymbol{x}}_i^k)\|^{2}                                                       \\
       & \leq(1+2\eta_{l}L)\mathbb{E}\|\boldsymbol{x}_{i}^{k}-\tilde{\boldsymbol{x}}_{i}^{k}\|^{2}+\frac{4\eta_{l}^{2}\sigma_{i}^{2}}{b_i} + 12\eta_{l}^{2}\zeta_i^2 + 6\eta_{l}^{2}L^2\mathbb{E}\|\boldsymbol{x}_{i}^{k}-\tilde{\boldsymbol{x}}_{i}^{k}\|^{2}                                                                               \\
       & = (1+2\eta_{l}L+6\eta_{l}^{2}L^2)\mathbb{E}\|\boldsymbol{x}_{i}^{k}-\tilde{\boldsymbol{x}}_{i}^{k}\|^{2}+\frac{4\eta_{l}^{2}\sigma_{i}^{2}}{b_i} + 12\eta_{l}^{2}\zeta_i^2
    \end{align*}

    \textbf{Bounding local uniform Stability:} Now, we can combine these two cases. Our bound relies on the probability of whether the perturbed samples are involved. Thus, we have:
    \begin{align*}
       & \mathbb{E}_{\mathcal{D}_i,\mathcal{D}_i'}\|\boldsymbol{x}_{i}^{k+1}-\tilde{\boldsymbol{x}}_{i}^{k+1}\|^{2}                                                                                                                                            \\
       & = P(\tilde{z}\notin\xi)\cdot \mathbb{E}[\|\boldsymbol{x}_{i}^{k+1}-\tilde{\boldsymbol{x}}_{i}^{k+1}\|^{2}|\tilde{z}\notin\xi] +  P(\tilde{z}\in\xi)\cdot\mathbb{E}[\|\boldsymbol{x}_{i}^{k+1}-\tilde{\boldsymbol{x}}_{i}^{k+1}\|^{2}|\tilde{z}\in\xi] \\
       & \leq (1-\frac{b_i}{n_i})\left((1+2\eta_{l}L+2\eta_{l}^{2}L^{2})\mathbb{E}\|\boldsymbol{x}_{i}^{k}-\tilde{\boldsymbol{x}}_{i}^{k}\|^{2}+\frac{4\eta_{l}^{2}\sigma_{i}^{2}}{b_i}\right)                                                                 \\
       & \quad + \frac{b_i}{n_i}\left((1+2\eta_{l}L+6\eta_{l}^{2}L^2)\mathbb{E}\|\boldsymbol{x}_{i}^{k}-\tilde{\boldsymbol{x}}_{i}^{k}\|^{2}+4\eta_{l}^{2}(\frac{\sigma_i^2}{b_i} + 3\zeta_i^2)\right)                                                         \\
       & = (1+2\eta_lL+2(1+2b_i/n_i)\eta_l^2L^2)\mathbb{E}\|\boldsymbol{x}_i^{k}-\tilde{\boldsymbol{x}}_i^{k}\|^2+4(\frac{\sigma_i^2}{n_i}+\frac{3b_i\zeta_i^2}{n_i})\eta_l^2\quad\quad\triangleright b_i/n_i\leq1/2                                           \\
       & \leq (1+2\eta_lL+4\eta_l^2L^2)\mathbb{E}\|\boldsymbol{x}_i^{k}-\tilde{\boldsymbol{x}}_i^{k}\|^2+4(\frac{\sigma_i^2}{n_i}+\frac{3b_i\zeta_i^2}{n_i})\eta_l^2\quad\quad\triangleright1+2\eta_lL\leq2                                                    \\
       & \leq (1+4\eta_lL)\mathbb{E}\|\boldsymbol{x}_i^{k}-\tilde{\boldsymbol{x}}_i^{k}\|^2+4(\frac{\sigma_i^2}{n_i}+\frac{3b_i\zeta_i^2}{n_i})\eta_l^2,
    \end{align*}
    where $\xi$ denotes the mini-batch data sampled from the local dataset. According to the model merge rule, the models fine-tuned for different tasks are all trained from the same pre-trained model, i.e., $\boldsymbol{x}_i^0 = \boldsymbol{x}^0$. Then, we unroll the above equation over local steps from $K_i - 1$ down to $k = 0$ $(K_i > 1):$
    \begin{align*}
    \mathbb{E}_{\mathcal{D}_i,\mathcal{D}_i^{\prime}}\|\boldsymbol{x}_i^{K_i}-\tilde{\boldsymbol{x}}_i^{K_i}\|^2 & \leq(1+4\eta_lL)^{K_i}\mathbb{E}\|\boldsymbol{x}^0-\tilde{\boldsymbol{x}}^0\|^2+\frac{(1+4\eta_lL)^{K_i}-1}{4\eta_lL}\cdot4\eta_l^2(\frac{\sigma_i^2}{n_i}+\frac{3b_i\zeta_i^2}{n_i}) \\                                                               & = \frac{(1+4\eta_lL)^{K_i}-1}{4\eta_lL}\cdot4\eta_l^2(\frac{\sigma_i^2}{n_i}+\frac{3b_i\zeta_i^2}{n_i})
    \end{align*}
    Here we use $\boldsymbol{x}^0 = \tilde{\boldsymbol{x}}^0$, which holds because both the perturbed and the unperturbed parameters are fine-tuned from the same pre-trained model.
    Assume the learning rate satisfies $\frac{1}{8KL} \geq \eta_l$, then we have $(1 + 4\eta_l L)^{K_i} - 1 \leq 4\eta_l LK_i$. We get:
    \begin{align*}
      \mathbb{E}_{\mathcal{D}_i,\mathcal{D}_i^{\prime}}\|\boldsymbol{x}_i^{K_i}-\tilde{\boldsymbol{x}}_i^{K_i}\|^2 \leq 16K_i\eta_l^2\left(\frac{\sigma_i^2}{n_i}+\frac{3b_i\zeta_i^2}{n_i}\right)
    \end{align*}
    Then we complete the proof.
  \end{proof}
\end{lemma}

\textbf{Step 2: Proofs of Global Stability}

Since model merging involves only a single aggregation step, different merging methods such as Task Arithmetic \cite{TA2022editing} and Adamerging \cite{yang2023adamerging} essentially learn different merging coefficients. Since higher-order model merging methods such as AdaMerging are capable of automatically capturing the differences between tasks, they can adaptively adjust the merging coefficients to achieve better performance.
This is equivalent to modifying the objective function from the standard form $f(\boldsymbol{x}) = \frac{1}{N} \sum_{i=1}^N f_i(\boldsymbol{x})$ to a weighted form $f(\boldsymbol{x}) = \sum_{i=1}^N \lambda_i f_i(\boldsymbol{x})$.
We will focus on discussing the objective function discrepancies caused by heterogeneity in tep 3.
The key difference among various model merging approaches lies in how they learn the parameters $\lambda_i$. To generalize our theoretical results, we will adopt the more general loss function $f(\boldsymbol{x}) = \sum_{i=1}^N \lambda_i f_i(\boldsymbol{x})$ in the analysis of Global Stability, where $\sum_{i=1}^N \lambda_i = 1$ according to Assumption \ref{ass:coeff}.

Specifically, the model merge step can be formally expressed as $\boldsymbol{x}_{avg} = \sum_{i=1}^N \lambda_i \boldsymbol{x}_i^{K_i}$, $\tilde{\boldsymbol{x}}_{avg} = \sum_{i=1}^N \lambda_i \tilde{\boldsymbol{x}}_i^{K_i}$. Then, the Global Stability can be represented as $\mathbb{E}_{\mathcal{D}, \mathcal{D}^{(j)}} \| \boldsymbol{x}_i^{K_i} - \tilde{\boldsymbol{x}}_i^{K_i} \|^2$. Next, we derive an upper bound for the Global Stability.
\begin{align*}
   & \mathbb{E}_{\mathcal{D}, \mathcal{D}^{(j)}} \| \boldsymbol{x}_{avg} - \tilde{\boldsymbol{x}}_{avg} \|^2                                                                                                                                                                                                                                                                                              \\
   & = \mathbb{E}_{\mathcal{D}, \mathcal{D}^{(j)}} \| \sum_{i=1}^N \lambda_i (\boldsymbol{x}_i^{K_i} - \tilde{\boldsymbol{x}}_i^{K_i}) \|^2 \quad\quad\triangleright \text{Jensen’s Inequality}                                                                                                                                                                                                           \\
   & \leq \sum_{i=1}^N\lambda_i\mathbb{E}_{\mathcal{D}, \mathcal{D}^{(j)}} \|\boldsymbol{x}_i^{K_i} - \tilde{\boldsymbol{x}}_i^{K_i}\|^2                                                                                                                                                                                                                                                                  \\
   & =\sum_{i=1}^N\lambda_i\left(\mathcal{P}(\tilde{z}\in\mathcal{D}_i)\cdot\mathbb{E}\left[\|\boldsymbol{x}_i^{K_i}-\tilde{\boldsymbol{x}}_i^{K_i}\|^2|\tilde{z}\in\mathcal{D}_i\right]+\mathcal{P}(\tilde{z}\notin\mathcal{D}_i)\cdot\mathbb{E}\left[\|\boldsymbol{x}_i^{K_i}-\tilde{\boldsymbol{x}}_i^{K_i}\|^2|\tilde{z}\notin\mathcal{D}_i\right]\right)                                             \\
   & = \sum_{i=1}^N\lambda_i\left(\frac{n_i}{n}\cdot\underbrace{\mathbb{E}\left[\|\boldsymbol{x}_i^{K_i}-\tilde{\boldsymbol{x}}_i^{K_i}\|^2|\tilde{z}\in\mathcal{D}_i\right]}_{\text{Lemma \ref{lemma2:local stability}}}+ (1-\frac{n_i}{n})\cdot\underbrace{\mathbb{E}\left[\|\boldsymbol{x}_i^{K_i}-\tilde{\boldsymbol{x}}_i^{K_i}\|^2|\tilde{z}\notin\mathcal{D}_i\right]}_{\text{Given Below}}\right) \\
\end{align*}

Then, we let $b_i=0$ for (\ref{eq:local stability}) denote the local stability without the perturbed sample (i.e., $\tilde{z}\notin\mathcal{D}_i).$ In this case, the local mini-batch SGD updates are amplified due to local stochastic gradient variance and cumulative SGD steps as:
\begin{align*}
  \mathbb{E}\left[\|\boldsymbol{x}_i^{K_i}-\tilde{\boldsymbol{x}}_i^{K_i}\|^2|\tilde{z}\notin\mathcal{D}_i\right] \leq 16K_i\eta_l^2\frac{\sigma_i^2}{n_i}
\end{align*}
Combining the above equations, we have
\begin{align}\label{eq:stability}
  \mathbb{E}_{\mathcal{D}, \mathcal{D}^{(j)}} \| \boldsymbol{x}_{avg} - \tilde{\boldsymbol{x}}_{avg} \|^2
  \leq 16\eta_l^2\sum_{i=1}^N\lambda_iK_i\left(\frac{\sigma_i^2}{n_i} + \frac{3b_i}{n_i}\zeta_i^2\right)
\end{align}
This concludes the proof of global stability.

\textbf{Step 3: Convergence Analysis (Optimization Error)}


Due to the heterogeneity in training resources or algorithmic hyperparameters during the fine-tuning stage across different tasks (e.g., different tasks may use different learning rates $\eta_l$, epochs $K_i$, optimizers, or batch sizes $b_i$), there may exist significant discrepancies among the parameters learned by each task. Whether merging such diverse model parameters can still lead to convergence is an important question worth exploring. To address this, we follow the analytical framework proposed by Wang et al. \cite{wang2020tackling}, with necessary modifications to suit the context of model merging. We will first introduce some notation and then analyze the theoretical implications of such heterogeneity.

\textbf{Notations:}
First, we define the following matrix of stochastic gradients for each task $i$:
\begin{align*}
  \boldsymbol{G}_i = [\boldsymbol{g}_i(\boldsymbol{x}_i^0), \boldsymbol{g}_i(\boldsymbol{x}_i^1),\cdots,\boldsymbol{g}_i(\boldsymbol{x}_i^{K_i-1})] \in \mathbb{R}^{d\times K_i}
\end{align*}
Here, $\boldsymbol{g}_i$ denotes a stochastic gradient of $f_i(\boldsymbol{x})$. Next, we define the normalized learning rate vector.
\begin{align*}
  \boldsymbol{a}_i = \left[\frac{\eta_i^0}{\eta_l},\frac{\eta_i^1}{\eta_l},\cdots, \frac{\eta_i^{K_i-1}}{\eta_l} \right]^{T} \in \mathbb{R}^{K_i}
\end{align*}
Here, $\eta_l$ denotes a constant learning rate, while $\eta_i^t$, with $t \in [K_i],\ i \in [N]$, can represent learning rates under any strategy, such as exponential decay.
\begin{equation}\label{eq:normal}
  \begin{aligned}
    \boldsymbol{x}_{avg} & = \boldsymbol{x}_0 + \frac{1}{N}\sum_{i=1}^N(\boldsymbol{x}_{i}^{K_i} - \boldsymbol{x}_0)  \\
    & = \boldsymbol{x}_0 - \frac{1}{N}\sum_{i=1}^N\eta_l \boldsymbol{G}_i \boldsymbol{a}_i  \\
    & = \boldsymbol{x}_0 - \underbrace{\left(\frac{1}{N}\sum_{i=1}^N\|\boldsymbol{a}_i\|_1\right)}_{\tau_{\text{eff}}:\;\text{effective local steps}}\sum_{i=1}^N\eta_l \underbrace{\left(\frac{\frac{1}{N}\|\boldsymbol{a}_i\|_1}{\frac{1}{N}\sum_{i=1}^N\|\boldsymbol{a}_i\|_1}\right)}_{\lambda_i:\; weight} \underbrace{\left(\frac{\boldsymbol{G}_i \boldsymbol{a}_i}{\|\boldsymbol{a}_i\|_1}\right)}_{\boldsymbol{d}_i:\; \text{normalized gradient}} \\
  \end{aligned}
\end{equation}
\textbf{Impact of Heterogeneity:} Unlike homogeneous training—where each task uses the same learning rate, number of epochs, batch size, etc.—heterogeneous training results in parameter updates that differ across tasks. According to the mathematical implication of Equation (\ref{eq:normal}), such heterogeneity effectively transforms the original optimization objective $f(\boldsymbol{x}) = \frac{1}{N} \sum_{i=1}^N f_i(\boldsymbol{x})$ into a surrogate objective $\tilde{f}(\boldsymbol{x}) = \sum_{i=1}^N \lambda_i f_i(\boldsymbol{x})$, where the coefficients $\lambda_i$ are defined as in Equation (\ref{eq:normal}).
Therefore, training heterogeneity introduces inconsistency in the objective function, making the theoretical analysis of model merging under heterogeneous conditions more complex and also more meaningful.
In the following analysis, we first present a more general theorem, and then provide a simplified version that establishes an upper bound on the optimization error under the assumption of an SGD optimizer. In the following analysis, we first present a more general theorem, which is adapted from the work of \cite{wang2020tackling}. However, we have made several modifications to their original result, as our Assumption \ref{ass:bound_hetero} differs from Assumption 3 in \cite{wang2020tackling}. Nevertheless, we show that our assumption implies theirs, and the proof is provided as follows:
\begin{assumption}\label{ass:in wang etal}
  (Bounded Dissimilarity in \cite{wang2020tackling})
  Let the merging coefficients satisfy Assumption \ref{ass:coeff}, and suppose there exist constants $\beta^2 \leq 1$ and $\kappa^2 \leq 1$ such that
  \[
    \sum_{i=1}^N \lambda_i \|\nabla f_i(\boldsymbol{x})\|^2 \leq \beta^2 \left\| \sum_{i=1}^N \lambda_i \nabla f_i(\boldsymbol{x}) \right\|^2 + \kappa^2.
  \]
\end{assumption}

\begin{lemma}[\textbf{Rigorous Satisfaction of Bounded Dissimilarity}]\label{lemma:beta_fix}
Under Assumption \ref{ass:bound_hetero} (Bounded Heterogeneity) and Assumption \ref{ass:coeff} (Coefficients $\sum \lambda_i=1$), the surrogate objective strictly satisfies the Bounded Dissimilarity condition required by existing convergence frameworks (e.g., Wang et al. \cite{wang2020tackling}) with constants $\beta^2 = 1 \leq 1$ and $\kappa^2 = \sum_{i=1}^N \lambda_i \zeta_i^2$.
\end{lemma}

\begin{proof}
Previous analyses often rely on loose algebraic inequalities such as $\|\boldsymbol{a}+\boldsymbol{b}\|^2 \leq 2\|\boldsymbol{a}\|^2 + 2\|\boldsymbol{b}\|^2$, which erroneously yields a loose constant $\beta^2=2$. Instead, we achieve a tight bound by leveraging the exact variance decomposition identity for weighted vectors.

By definition, the global aggregated gradient is the $\boldsymbol{\lambda}$-weighted mean of the local gradients: $\nabla f(\boldsymbol{x}) = \sum_{i=1}^N \lambda_i \nabla f_i(\boldsymbol{x})$. 

We evaluate the weighted variance of the local gradients around their global mean:
\begin{equation}
\begin{aligned}
& \sum_{i=1}^N \lambda_i \|\nabla f_i(\boldsymbol{x}) - \nabla f(\boldsymbol{x})\|^2 \\
&= \sum_{i=1}^N \lambda_i \left( \|\nabla f_i(\boldsymbol{x})\|^2 - 2 \langle \nabla f_i(\boldsymbol{x}), \nabla f(\boldsymbol{x}) \rangle + \|\nabla f(\boldsymbol{x})\|^2 \right) \\
&= \sum_{i=1}^N \lambda_i \|\nabla f_i(\boldsymbol{x})\|^2 - 2 \left\langle \sum_{i=1}^N \lambda_i \nabla f_i(\boldsymbol{x}), \nabla f(\boldsymbol{x}) \right\rangle + \left( \sum_{i=1}^N \lambda_i \right) \|\nabla f(\boldsymbol{x})\|^2
\end{aligned}
\end{equation}

Substituting $\sum_{i=1}^N \lambda_i \nabla f_i(\boldsymbol{x}) = \nabla f(\boldsymbol{x})$ and $\sum_{i=1}^N \lambda_i = 1$ into the equation yields:
\begin{equation}
\begin{aligned}
\sum_{i=1}^N \lambda_i \|\nabla f_i(\boldsymbol{x}) - \nabla f(\boldsymbol{x})\|^2 &= \sum_{i=1}^N \lambda_i \|\nabla f_i(\boldsymbol{x})\|^2 - 2 \langle \nabla f(\boldsymbol{x}), \nabla f(\boldsymbol{x}) \rangle + 1 \cdot \|\nabla f(\boldsymbol{x})\|^2 \\
&= \sum_{i=1}^N \lambda_i \|\nabla f_i(\boldsymbol{x})\|^2 - 2 \|\nabla f(\boldsymbol{x})\|^2 + \|\nabla f(\boldsymbol{x})\|^2 \\
&= \sum_{i=1}^N \lambda_i \|\nabla f_i(\boldsymbol{x})\|^2 - \|\nabla f(\boldsymbol{x})\|^2
\end{aligned}
\end{equation}

By simply rearranging the terms, we obtain the fundamental identity:
\begin{equation}\label{eq:variance_identity}
\sum_{i=1}^N \lambda_i \|\nabla f_i(\boldsymbol{x})\|^2 = \|\nabla f(\boldsymbol{x})\|^2 + \sum_{i=1}^N \lambda_i \|\nabla f_i(\boldsymbol{x}) - \nabla f(\boldsymbol{x})\|^2
\end{equation}

Taking the expectation on both sides and applying Assumption \ref{ass:bound_hetero} ($\mathbb{E}\|\nabla f_i(\boldsymbol{x}) - \nabla f(\boldsymbol{x})\|^2 \leq \zeta_i^2$), we get:
\begin{equation}
\mathbb{E} \left[ \sum_{i=1}^N \lambda_i \|\nabla f_i(\boldsymbol{x})\|^2 \right] \leq 1 \cdot \mathbb{E} \|\nabla f(\boldsymbol{x})\|^2 + \sum_{i=1}^N \lambda_i \zeta_i^2
\end{equation}

This perfectly aligns with the Bounded Dissimilarity requirement $\mathbb{E}\sum \lambda_i \|\nabla f_i\|^2 \leq \beta^2 \mathbb{E}\|\nabla f\|^2 + \kappa^2$, rigorously verifying that $\beta^2 = 1 \leq 1$ and $\kappa^2 = \sum_{i=1}^N \lambda_i \zeta_i^2$. 
\end{proof}

\begin{theorem}\label{the:wang_etal}
  (Convergence to the Surrogate Objective $\tilde{f}(\boldsymbol{x})$ Stationary Point from \cite{wang2020tackling})
  Under Assumptions \ref{ass:smooth}, \ref{ass:bound SG}, and \ref{ass:in wang etal}, let $\bar{K} = \frac{1}{N} \sum_{i=1}^N K_i$ and $\eta_l = \sqrt{\frac{N}{\bar{K}}}$. Then, the gradient of the surrogate function $\tilde{f}(\boldsymbol{x}) = \sum_{i=1}^N \lambda_i f_i(\boldsymbol{x})$ at the averaged model $\boldsymbol{x}_{avg}$ is bounded as follows:
  \begin{equation}\label{eq:the1}
    \begin{aligned}
      \mathbb{E}[\|\nabla\tilde{f}(\boldsymbol{x}_{avg})\|^2] \leq \frac{4(\tilde{f}(\boldsymbol{x}_0) - \tilde{f}_{\inf})(\bar{K}/\tau_{eff})}{\sqrt{N\bar{K}}}+\frac{4L\sigma^2A_1}{b\sqrt{N\bar{K}}}+\frac{6NL^2\sigma^2A_2}{b\bar{K}}+\frac{12NL^2\zeta^2A_3}{\bar{K}}.
    \end{aligned}
  \end{equation}

  Where $\tilde{f}_{\inf} = \underset{\boldsymbol{x}}{\min} \tilde{f}(\boldsymbol{x}), \sigma = \underset{i}{\max}{\sigma_i}, b =\underset{i}{\min}{ b_i},A_{1}=\tau_{eff}N\sum_{i=1}^{N}\frac{\lambda_{i}^{2}\|\boldsymbol{a}_{i}\|_{2}^{2}}{\|\boldsymbol{a}_{i}\|_{1}^{2}}, A_2=\sum_{i=1}^N\lambda_i(\|\boldsymbol{a}_i\|_2^2-a_{i,-1}^2)$
  and $A_3 = \max_i\{\|\boldsymbol{a}_i\|_1(\|\boldsymbol{a}_i\|_1-a_{i,-1})\}$, where $ a_{i,-1}$ denotes the last coordinate of the vector $\boldsymbol{a}_t$.
  Furthermore, let the right-hand side of Equation \ref{eq:the1} be denoted as $\epsilon$. Then, we have the following inequality, which ensures that even under heterogeneous conditions, the original objective function $f(\boldsymbol{x})$ can still converge, although its bound is not as tight as that in Equation \ref{eq:the1}:
  \begin{equation}
    \mathbb{E}[\|\nabla f(\boldsymbol{x}_{avg})\|^2] \leq 2[\chi_{\boldsymbol{p}||\boldsymbol{\lambda}}^2 + 1]\epsilon + 2\chi_{\boldsymbol{p}||\boldsymbol{\lambda}}^2\sum_{i=1}^N\lambda_i\zeta_i^2
  \end{equation}
  where $\chi_{\boldsymbol{p}||\boldsymbol{\lambda}}^2 = \sum_{i=1}^N\frac{(\frac{1}{N} - \lambda_i)^2}{\lambda_i^2}$ is the chi-square divergence between the weight coefficient vectors $\boldsymbol{\lambda} = [\lambda_1,\lambda_2,\cdots,\lambda_N]\in \mathbb{R}^N$ and $\boldsymbol{p} = [\frac{1}{N},\frac{1}{N},\cdots,\frac{1}{N} ]\in \mathbb{R}^N$.
\end{theorem}
In fact, different choices of optimizers in Theorem~\ref{the:wang_etal} yield different forms of the weighting vector $\boldsymbol{a}_i$. In the work of Wang et al.~\cite{wang2020tackling}, several illustrative examples are provided. For instance, when the optimizer incorporates a proximal term, the corresponding $\boldsymbol{a}_i$ is given by
\[
  \boldsymbol{a}_i = \big[(1-\alpha)^{K_i-1}, (1-\alpha)^{K_i-2}, \dots, (1-\alpha), 1\big] \in \mathbb{R}^{K_i},
\]
where $\alpha = \alpha \mu$ and $\mu$ is a tunable parameter. In this case, the effective number of iterations and the associated weights are defined as
\[
  \tau_{\text{eff}} = \sum_{i=1}^{N} \frac{1 - (1-\alpha)^{K_i}}{\alpha N}, \quad \lambda_i = \frac{1 - (1-\alpha)^{K_i}/N}{\sum_{i=1}^{N} \left[1 - (1-\alpha)^{K_i}\right]/N}.
\]
Moreover, Wang et al.~\cite{wang2020tackling} also derive explicit forms of $\boldsymbol{a}_i$, $\tau_{\text{eff}}$, and $\lambda_i$ when using other optimizers such as SGD with learning rate decay or SGD with momentum.

In summary, Theorem~\ref{the:wang_etal} provides a general analytical framework accommodating a variety of optimization strategies. For the sake of simplicity and to align with the hyperparameter settings used in the \textit{Step 2 Global Stability Analysis}, we proceed with a simplified version assuming that each task is optimized using standard SGD. Under this setting, we have
\[
  \boldsymbol{a}_i = [1,1,\dots,1] \in \mathbb{R}^{K_i}, \quad \tau_{\text{eff}} = \bar{K} = \frac{1}{N} \sum_{i=1}^{N} K_i, \quad \|\boldsymbol{a}_i\|_1 = K_i, \quad \lambda_i = \frac{K_i}{N \bar{K}},
\]
which leads directly to the following corollary of Theorem~\ref{the:wang_etal}.

\begin{corollary}
  (Adopt the same SGD optimizer for all tasks from Theorem \ref{the:wang_etal}) Under Assumptions \ref{ass:smooth}, \ref{ass:bound SG}, and \ref{ass:in wang etal}, let $\bar{K} = \frac{1}{N} \sum_{i=1}^N K_i$, $\eta_l = \sqrt{\frac{N}{\bar{K}}}$ and all task use the SGD optimizer. Then, the gradient of the surrogate function $\tilde{f}(\boldsymbol{x}) = \sum_{i=1}^N \lambda_i f_i(\boldsymbol{x})$ at the averaged model $\boldsymbol{x}_{avg}$ is bounded as follows:
  \begin{equation}\label{eq:corollary1}
    \begin{aligned}
      \mathbb{E}[\|\nabla\tilde{f}(\boldsymbol{x}_{avg})\|^2] \leq \frac{4(\tilde{f}(\boldsymbol{x}_0) - \tilde{f}_{\inf})}{\sqrt{N\bar{K}}}+\frac{4L\sigma^2A_1}{b\sqrt{N\bar{K}}}+\frac{6NL^2\sigma^2A_2}{b\bar{K}}+\frac{12NL^2\zeta^2A_3}{\bar{K}}.
    \end{aligned}
  \end{equation}

  Where $\tilde{f}_{\inf} = \underset{\boldsymbol{x}}{\min} \tilde{f}(\boldsymbol{x}), \sigma = \underset{i}{\max}{\{\sigma_i\}},b =\underset{i}{\min}{ b_i}, \zeta = \underset{i}{\max}{\{\zeta_i\}}, A_{1}=N\sum_{i=1}^{N}\frac{\bar{K}}{K_i}\lambda_{i}^{2}, A_2=\sum_{i=1}^N\lambda_i(K_i - 1)$
  and $A_3 = \max_i\{K_i(K_i - 1)\}$.
  Furthermore, let the right-hand side of Equation \ref{eq:corollary1} be denoted as $\epsilon_{sgd}$. Then, we have the following inequality, which ensures that even under heterogeneous conditions, the original objective function $f(\boldsymbol{x})$ can still converge, although its bound is not as tight as that in Equation \ref{eq:corollary1}:
  \begin{equation}\label{eq:normal gradient}
    \mathbb{E}[\|\nabla f(\boldsymbol{x}_{avg})\|^2] \leq 2[\chi_{\boldsymbol{p}||\boldsymbol{\lambda}}^2 + 1]\epsilon_{sgd} + 2\chi_{\boldsymbol{p}||\boldsymbol{\lambda}}^2\sum_{i=1}^N\lambda_i\zeta_i^2
  \end{equation}
  where $\chi_{\boldsymbol{p}||\boldsymbol{\lambda}}^2 = \sum_{i=1}^N\frac{(\frac{1}{N} - \lambda_i)^2}{\lambda_i^2}$ is the chi-square divergence between the weight coefficient vectors $\boldsymbol{\lambda} = [\lambda_1,\lambda_2,\cdots,\lambda_N]\in \mathbb{R}^N$ and $\boldsymbol{p} = [\frac{1}{N},\frac{1}{N},\cdots,\frac{1}{N} ]\in \mathbb{R}^N$.
\end{corollary}

\textbf{Step 4: Excess Risk Analysis via Joint Minimization}

According to the definition of excess error in Equation~\ref{eq:excess error}, we have $\mathcal{E}(\boldsymbol{x}) = \mathcal{E}_{G}(\boldsymbol{x}) + \mathcal{E}_{O}(\boldsymbol{x})$. Furthermore, we assume that the optimization error satisfies $\mathcal{E}_{O}(\boldsymbol{x}_{avg}) \leq C \cdot \mathbb{E}\|\nabla f(\boldsymbol{x}_{avg})\|^2$. When $C = \frac{1}{2\mu}$, this assumption degenerates to the Polyak-Łojasiewicz (PL) condition proposed in \cite{jain2017non}. However, for non-convex optimization analysis, the PL condition is sometimes considered too strong. A milder assumption is the Kurdyka-Łojasiewicz (KL) condition, which yields $\mathcal{E}_{O}(\boldsymbol{x}_{avg}) \leq C^{\frac{1}{2\theta}} \cdot \mathbb{E}\|\nabla f(\boldsymbol{x}_{avg})\|^{\frac{1}{\theta}}$, where $f$ is assumed to be smooth and $\theta \in [0,1)$. Notably, when $\theta = \frac{1}{2}$, the KL condition recovers the PL condition. To maintain generality, we do not explore how the KL parameter $\theta$ improves convergence \cite{karimi2016linear} or generalization \cite{charles2018stability}, thereby avoiding strong assumptions.

Combining Lemma~\ref{lemma:ob-avg-gen}, Equation~\ref{eq:excess error}, and the assumption $\mathcal{E}_{O}(\boldsymbol{x}_{avg}) \leq C \cdot \mathbb{E}\|\nabla f(\boldsymbol{x}_{avg})\|^2$, we derive the following stability-based expression for the excess error:
\begin{align*}
  \mathcal{E}(\boldsymbol{x}_{avg}) \leq \frac{L + \gamma}{2} \underbrace{\mathbb{E}\|\boldsymbol{x}_{avg} - \tilde{\boldsymbol{x}}_{avg}\|^2}_{\text{Model stability}} + \left(\frac{1}{2\gamma} + C\right) \underbrace{\mathbb{E}\|\nabla f(\boldsymbol{x}_{avg})\|^2}_{\text{Gradient norm}}.
\end{align*}

Substituting inequality \ref{eq:stability} and \ref{eq:corollary1} into the above equation, we obtain:

\begin{align}
  \mathcal{E}(\boldsymbol{x}_{avg}) \leq 8(L + \gamma)\eta_l^2\sum_{i=1}^N\lambda_iK_i\left(\frac{\sigma_i^2}{n_i} + \frac{3b_i}{n_i}\zeta_i^2\right)  + \left(\frac{1}{\gamma} + 2C\right) \left[\chi_{\boldsymbol{p}||\boldsymbol{\lambda}}^2\sum_{i=1}^N\lambda_i\zeta_i^2  + (\chi_{\boldsymbol{p}||\boldsymbol{\lambda}}^2 + 1)\epsilon_{sgd}\right]
\end{align}

Where $\epsilon_{sgd}$ is defined in inequality (\ref{eq:normal gradient}).

Next, we will determine the optimal $\gamma$ for the above equation to minimize the upper bound on the right-hand side.
After a straightforward differentiation, we obtain that when $\gamma$ takes the following form,
\begin{equation*}
  \gamma^* = \sqrt{\frac{\chi_{\boldsymbol{p}||\boldsymbol{\lambda}}^2\sum_{i=1}^N\lambda_i\zeta_i^2 + (\chi_{\boldsymbol{p}||\boldsymbol{\lambda}}^2 + 1)\epsilon_{sgd}}{8\eta_l^2\sum_{i=1}^N\lambda_iK_i\left(\frac{\sigma_i^2}{n_i} + \frac{3b_i}{n_i}\zeta_i^2\right)}}
\end{equation*}

Consequently, the tightest bound of $\mathcal{E}(\boldsymbol{x}_{avg})$ with respect to $\gamma$ is:

\begin{align*}
  \mathcal{E}(\boldsymbol{x}_{avg}) \leq & 8L\eta_l^2\sum_{i=1}^N\lambda_iK_i\left(\frac{\sigma_i^2}{n_i} + \frac{3b_i}{n_i}\zeta_i^2\right) + 2C\left[\chi_{\boldsymbol{p}||\boldsymbol{\lambda}}^2\sum_{i=1}^N\lambda_i\zeta_i^2 + (\chi_{\boldsymbol{p}||\boldsymbol{\lambda}}^2 + 1)\epsilon_{sgd}\right]                 \\
 & + 4\sqrt{2}\eta_l\sqrt{\sum_{i=1}^N\lambda_iK_i\left(\frac{\sigma_i^2}{n_i} + \frac{3b_i}{n_i}\zeta_i^2\right) \cdot \left[\chi_{\boldsymbol{p}||\boldsymbol{\lambda}}^2\sum_{i=1}^N\lambda_i\zeta_i^2 + (\chi_{\boldsymbol{p}||\boldsymbol{\lambda}}^2 + 1)\epsilon_{sgd}\right]}
\end{align*}

To facilitate the analysis, we apply the basic inequality $2\sqrt{ab}\leq a+b$ to the radical term in the above expression, yielding:

\begin{align}\label{eq:tight_results}
  \mathcal{E}(\boldsymbol{x}_{avg}) \leq & 8(L+1)\eta_l^2\sum_{i=1}^N\lambda_iK_i\left(\frac{\sigma_i^2}{n_i} + \frac{3b_i}{n_i}\zeta_i^2\right)                                                              \\ \nonumber
  \quad +                                & (2C+1)\left[\chi_{\boldsymbol{p}||\boldsymbol{\lambda}}^2\sum_{i=1}^N\lambda_i\zeta_i^2 + (\chi_{\boldsymbol{p}||\boldsymbol{\lambda}}^2 + 1)\epsilon_{sgd}\right]
\end{align}

\subsubsection{Deeper Analysis of the Generalization Bound and Hyperparameters}\label{app:hyperparameter_jieshi}

While understanding how different merging methods correspond to choices of $\boldsymbol{\lambda}$ is crucial, a deeper analysis of the hyperparameters within the excess risk bound in Equation \ref{eq:tight_results} reveals fundamental trade-offs in the fine-tune-then-merge paradigm. We now dissect the role of key hyperparameters.

The final bound is driven by two main components: a \textit{stability term}, scaled by $8(L+1)\eta_l^2$, and an \textit{optimization error term}, scaled by $(2C+1)$.

$$
  \mathcal{E}(\boldsymbol{x}_{avg}) \leq \underbrace{8(L+1)\eta_l^2\sum_{i=1}^N\lambda_iK_i\left(\frac{\sigma_i^2}{n_i} + \frac{3b_i}{n_i}\zeta_i^2\right)}_{\text{Stability Component}} + \underbrace{(2C+1)\left[\chi_{\boldsymbol{p}||\boldsymbol{\lambda}}^2\sum_{i=1}^N\lambda_i\zeta_i^2 + (\chi_{\boldsymbol{p}||\boldsymbol{\lambda}}^2 + 1)\epsilon_{sgd}\right]}_{\text{Optimization Error Component}}
$$

Let's analyze the impact of each hyperparameter:

\begin{enumerate}[leftmargin=15pt]
  \item \textbf{Number of Local Fine-Tuning Steps ($K_i$):} The number of local epochs or steps is a critical parameter that embodies the core trade-off between optimization and generalization.
        \begin{itemize}
          \item \textbf{Mathematical Impact:} $K_i$ appears linearly in the numerator of the stability component, suggesting that more steps can worsen stability. Conversely, the optimization error term $\epsilon_{sgd}$ generally decreases with the average number of steps, $\bar{K}$, typically at a rate of $O(1/\sqrt{\bar{K}})$ or $O(1/\bar{K})$.
          \item \textbf{Interpretation (The Optimization-Generalization Trade-off):}
                \begin{itemize}
                  \item A \textbf{small $K_i$} keeps the stability term low. The model stays close to the pre-trained initialization ($\boldsymbol{x}_0$), and its final state is less sensitive to individual data points in its local dataset $\mathcal{D}_i$. However, a small $K_i$ leads to a large $\epsilon_{sgd}$, meaning the model is poorly optimized (under-fitting).
                  \item A \textbf{large $K_i$} reduces the optimization error $\epsilon_{sgd}$, pushing the model closer to the minimum of its local objective. However, this comes at the cost of increased instability. With more steps, the model has more capacity to over-specialize on $\mathcal{D}_i$, causing a small data perturbation to lead to a larger divergence in the final parameters $\boldsymbol{x}_i^{K_i}$, which harms generalization.
                \end{itemize}
          \item \textbf{Practical Insight:} This trade-off implies that an optimal number of fine-tuning steps exists. Depending on which factor—optimization error or instability—dominates, performance may first increase and then decrease (an inverted U-shape), or it could decrease monotonically if the stability cost is immediately too high. Overtraining leads to over-specialized models that are difficult to merge, a phenomenon our bound quantifies as a loss of algorithmic stability. Therefore, we recommend incorporating more metrics to guide the selection of the optimal $K_i$ in practical tasks. In general, it is advisable to employ multiple monitoring mechanisms to determine whether the optimization process has converged. Once convergence is detected, training should be stopped immediately, as continuing to train will increase the stability term, while the reduction in optimization error becomes marginal and outweighed by the degradation in stability.
        \end{itemize}

  \item \textbf{Batch Size ($b_i$):} The batch size exhibits a subtle, dual role whose practical effect is conditioned by other hyperparameters.
        \begin{itemize}
          \item \textbf{Mathematical Impact:} $b_i$ appears in the numerator of the heterogeneity-driven part of the stability term ($3b_i\zeta_i^2/n_i$), suggesting a larger batch size can harm stability. However, the entire stability term is scaled by $\eta_l^2$.
          \item \textbf{Interpretation (A Conditional Trade-off):}
                \begin{itemize}
                  \item In principle, $b_i$ presents a trade-off. A \textit{small $b_i$} keeps the term $3b_i\zeta_i^2/n_i$ small but can lead to noisy gradients that slow convergence, affecting the optimization error $\epsilon_{sgd}$.
                  \item A \textbf{large $b_i$} reduces the variance of stochastic gradients, typically improving the optimization error $\epsilon_{sgd}$. However, it linearly increases the stability cost term $3b_i\zeta_i^2/n_i$, making the algorithm more sensitive to data perturbations.
                  \item The \textbf{decisive factor} is the learning rate $\eta_l$. Since the stability cost is scaled by $\eta_l^2$, a small learning rate can heavily suppress this negative term, making the positive effect on optimization the dominant factor.
                \end{itemize}
          \item \textbf{Practical Insight:} Our bound provides a theoretical rationale for why the observed effect of batch size depends on the experimental setting. While a trade-off exists in theory, for fine-tuning regimes with small learning rates, the benefits of improved gradient estimation from larger batch sizes are expected to dominate, leading to monotonically better performance. Therefore, we recommend using a large batch size for fine-tuning in practical tasks whenever possible.
        \end{itemize}

  \item \textbf{Dataset Size ($n_i$):} The amount of data per task has a clear and unambiguously beneficial effect.
        \begin{itemize}
          \item \textbf{Mathematical Impact:} $n_i$ appears in the denominator of the entire stability term, scaling both the variance component ($\sigma_i^2/n_i$) and the heterogeneity component ($3b_i\zeta_i^2/n_i$).
          \item \textbf{Interpretation (A "Free Lunch" for Generalization):} Unlike other hyperparameters that present a trade-off, increasing $n_i$ strictly tightens the generalization bound by enhancing model stability without adversely affecting the optimization error component. The influence of any single data point is diminished as the dataset grows, reducing the model's sensitivity to data perturbations.
          \item \textbf{Practical Insight:} This aligns with the foundational principle of statistical learning: more data leads to better generalization. Our bound confirms this holds true in the context of model merging, predicting a direct and monotonic improvement in performance with more data. Therefore, we recommend using as much data as possible for fine-tuning in practical tasks.
        \end{itemize}

  \item \textbf{Learning Rate ($\eta_l$):} The learning rate is a dominant factor in controlling stability.
        \begin{itemize}
          \item \textbf{Mathematical Impact:} $\eta_l$ appears as a squared term, $\eta_l^2$, multiplying the entire stability component.
          \item \textbf{Interpretation (The Convergence vs. Stability Trade-off):}
                \begin{itemize}
                  \item A \textbf{small $\eta_l$} is essential for good generalization. The $\eta_l^2$ term indicates that the stability bound is highly sensitive to the learning rate. A smaller step size ensures that any perturbation in the gradient results in only a small change in the final parameters, thus promoting stability.
                  \item A \textbf{large $\eta_l$} can be catastrophic for generalization. It can cause the stability component to explode, leading to an unstable algorithm whose output varies wildly with small data changes, potentially even causing model collapse after merging.
                \end{itemize}
          \item \textbf{Practical Insight:} The bound provides strong theoretical justification for using small learning rates during fine-tuning. This is crucial not only to avoid catastrophic forgetting but also, as our theory shows, to ensure the resulting model is stable enough to be effectively merged and to generalize well. Therefore, we recommend using a small learning rate for fine-tuning in practical tasks.
        \end{itemize}

  \item \textbf{Number of Tasks ($N$):} The number of tasks being merged introduces a trade-off between a minor optimization benefit and a significant heterogeneity cost.
        \begin{itemize}
          \item \textbf{Mathematical Impact:} $N$ appears in the denominator of the optimization error term $\epsilon_{sgd}$ (improving at a slow rate of $O(1/\sqrt{N})$), but the total task heterogeneity, which impacts several terms, grows linearly with $N$.
          \item \textbf{Interpretation (The Optimization vs. Heterogeneity Imbalance):}
                \begin{itemize}
                  \item A \textbf{larger $N$} provides a small, slowly-improving benefit to the optimization error. However, this benefit's leading coefficient, related to the initial suboptimality gap, is small for models finetuned from strong pretrained checkpoints.
                  \item Conversely, as more tasks are added, the average task heterogeneity increases. This "parameter pollution" from conflicting tasks creates a significant penalty that accumulates rapidly. There is a fundamental imbalance: a minor benefit with a small coefficient is pitted against a substantial penalty that grows quickly.
                \end{itemize}
          \item \textbf{Practical Insight:} Our bound predicts that for a diverse set of tasks, the performance degradation from escalating task heterogeneity will overwhelm the marginal gains in optimization. This leads to the expectation that generalization performance will decrease as $N$ increases, highlighting a key limitation of simple averaging in many-task scenarios. Therefore, we recommend minimizing the inclusion of models unrelated to the target task during merging, as reducing the number of merged models can improve overall performance.
        \end{itemize}
\end{enumerate}

\section{Experiment Details}\label{app:exp_detail}

\subsection{Datasets Setup}

We evaluate our methods on a diverse set of 20 downstream tasks spanning multiple domains. These include natural scene classification (SUN397~\cite{xiaoSUNDatabaseLargescale2010}), fine-grained object recognition (Stanford-Cars~\cite{krause3DObjectRepresentations2013}, Oxford Flowers102~\cite{nilsback2008automated}), remote sensing (RESISC45~\cite{chengRemoteSensingImage2017}, EuroSAT~\cite{helberEuroSATNovelDataset2019}), digit and character recognition (MNIST~\cite{lecunGradientbasedLearningApplied1998}, SVHN~\cite{netzerReadingDigitsNatural2011}, EMNIST~\cite{cohen2017emnist}, KMNIST~\cite{clanuwat2018deep}), traffic sign recognition (GTSRB~\cite{stallkampManVsComputer2012}), texture classification (DTD~\cite{cimpoiDescribingTexturesWild2014}), medical imaging (PCAM~\cite{veeling2018rotation}), facial expression recognition (FER2013~\cite{goodfellow2013challenges}), pet breed classification (Oxford-IIIT-Pet~\cite{parkhi2012cats}), general object recognition (CIFAR10, CIFAR100~\cite{krizhevsky2009learning}, STL10~\cite{coates2011analysis}, Food101~\cite{bossard2014food}), fashion product classification (Fashion-MNIST~\cite{xiao2017fashion}), and sentiment analysis (Rendered-SST2~\cite{socher2013recursive,radford2021learning}). This selection covers a wide range of visual and textual tasks, ensuring comprehensive evaluation of model merging strategies under heterogeneous domains and data distributions.

\section{More Results}

\subsection{The Average Performance of Individual Models Before Merging}

\begin{figure*}[ht]
  \centering
  \includegraphics[width=\textwidth]{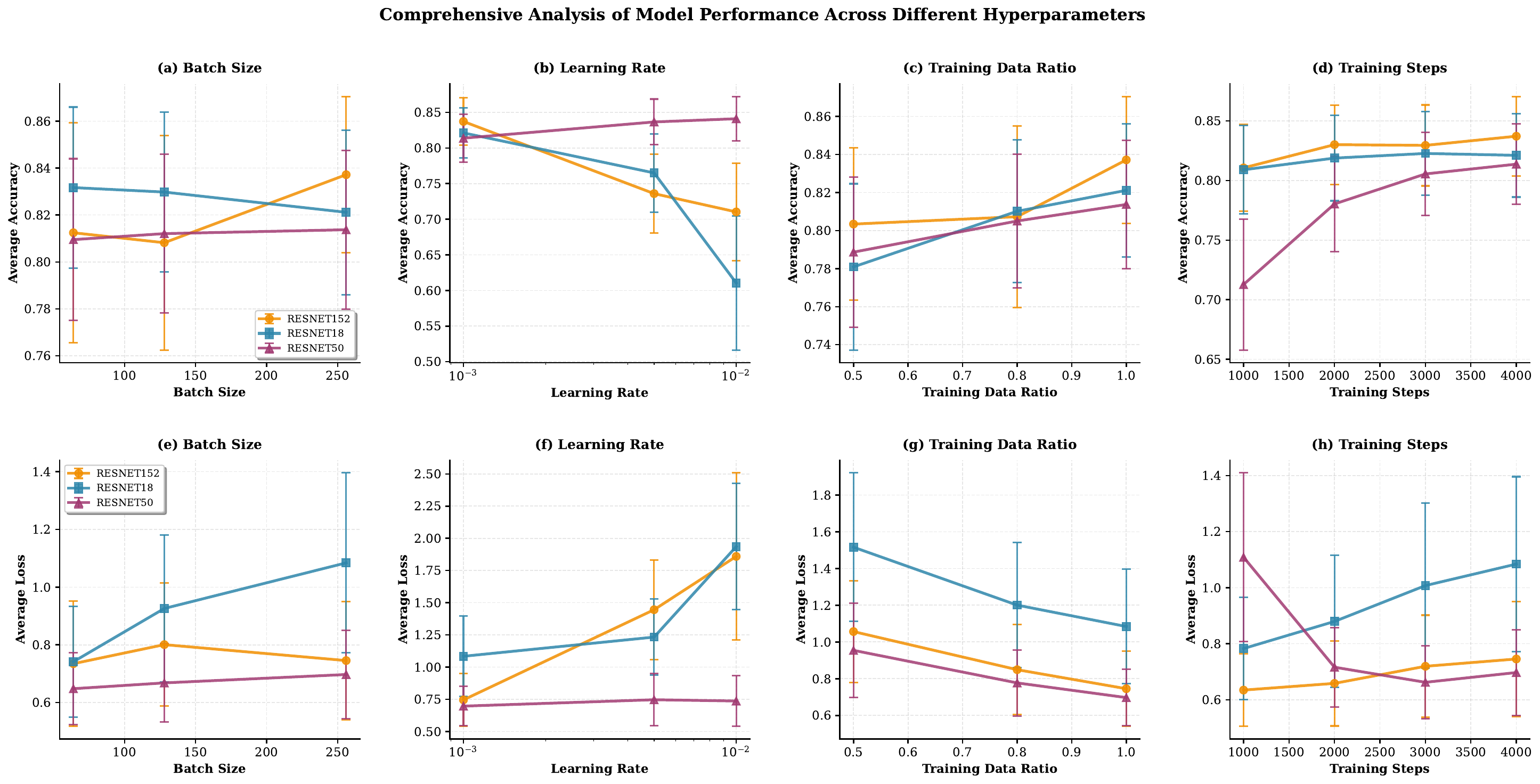}
  \caption{\textbf{Hyperparameter Impact on Model Merging Performance.} Comprehensive analysis of four key finetuning hyperparameters across three ResNet architectures. Top row (a-d): accuracy vs. batch size, learning rate, training data ratio, and training steps. Bottom row (e-h): corresponding loss curves. Each experiment varies one hyperparameter while fixing others at defaults (batch size=256, lr=0.001, data ratio=1.0, steps=4000). Results averaged over 20 vision datasets with error bars showing standard error.}
  \label{fig:main_results_single_finetune}
\end{figure*}

\subsection{Single-Task Finetuning Performance}
\label{sec:appendix_single_task}

To provide a comprehensive baseline for our merging experiments, the following table details the performance of each backbone model after being individually fine-tuned on its respective task. These accuracy scores represent the "expert" model performance before any model merging techniques are applied.

\begin{table}[h!]
\centering
\caption{\textbf{Single-Task Finetuning Accuracy.} Performance (top-1 accuracy) of ResNet-18, ResNet-50, and ResNet-152 models after being individually fine-tuned on each of the 20 downstream vision tasks. \textbf{All models were fine-tuned for 4000 steps with a learning rate of 0.001, a batch size of 256, and using the full training dataset (data ratio of 1.0).} The highest accuracy for each task is highlighted in \textbf{bold}. This data serves as the baseline for the expert models used in our merging experiments.}
\label{tab:appendix_single_task_perf}
\begin{tabular}{l c c c}
\toprule
\textbf{Task} & \textbf{ResNet-18} & \textbf{ResNet-50} & \textbf{ResNet-152} \\
\midrule
cifar10           & 0.9404 & 0.9480 & \textbf{0.9743} \\
cifar100          & 0.7683 & 0.7712 & \textbf{0.8435} \\
dtd               & 0.6676 & 0.6984 & \textbf{0.7223} \\
emnist\_letters   & \textbf{0.9485} & 0.9334 & 0.9470 \\
eurosat           & \textbf{0.9826} & 0.9711 & 0.9770 \\
fashion\_mnist    & \textbf{0.9353} & 0.9084 & 0.9282 \\
fer2013           & \textbf{0.6553} & 0.5853 & 0.6495 \\
food101           & 0.7303 & 0.7200 & \textbf{0.7763} \\
gtsrb             & \textbf{0.9594} & 0.9174 & 0.9478 \\
kmnist            & \textbf{0.9696} & 0.9188 & 0.9601 \\
mnist             & 0.9946 & 0.9878 & \textbf{0.9946} \\
oxford-iiit-pet   & 0.8613 & 0.9057 & \textbf{0.9201} \\
oxford\_flowers102& \textbf{0.7637} & 0.7536 & 0.7291 \\
pcam              & 0.8242 & 0.8357 & \textbf{0.8390} \\
rendered-sst2     & 0.5211 & \textbf{0.5409} & 0.5041 \\
resisc45          & \textbf{0.9063} & 0.8760 & 0.9030 \\
stanford-cars     & 0.6241 & 0.5769 & \textbf{0.6260} \\
stl10             & 0.9408 & 0.9668 & \textbf{0.9813} \\
sun397            & 0.4794 & 0.5329 & \textbf{0.5742} \\
svhn              & \textbf{0.9501} & 0.9262 & 0.9458 \\
\bottomrule
\end{tabular}
\end{table}

\clearpage
\subsection{Detailed Experimental Results}
\label{sec:appendix_results}

This section provides the complete set of visualizations for the experimental results discussed in Section~\ref{sec:verification}. Each figure corresponds to a specific hyperparameter analysis, illustrating the performance of the merged model across three different ResNet backbones (ResNet-18, ResNet-50, and ResNet-152). For each hyperparameter, we present both line plots showing trends with error bars (top row) and bar charts for direct comparison at specific values (bottom row), evaluated by both average accuracy and average loss.

\begin{figure}[h]
  \centering
  \includegraphics[width=0.8\textwidth]{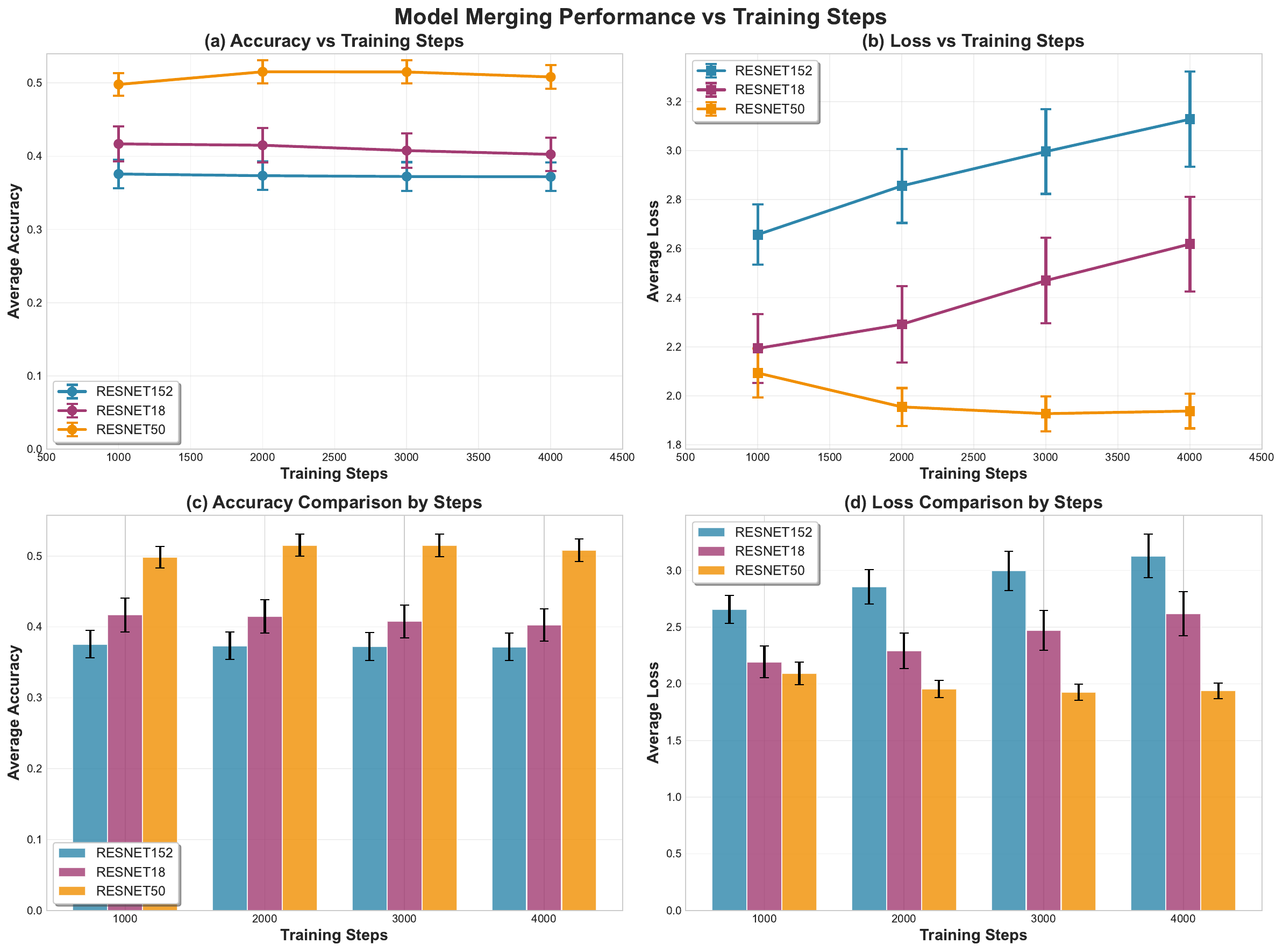} 
  \caption{\textbf{Detailed results for the impact of fine-tuning steps ($K_i$).} The top row (a, b) shows performance trends, while the bottom row (c, d) provides direct comparisons. The results exhibit a clear trade-off: performance initially improves with more steps but then degrades, particularly for ResNet-18 and ResNet-50. This inverted U-shaped trend for accuracy and U-shaped trend for loss strongly substantiates our theoretical prediction of a balance between optimization and generalization, as discussed in Section~\ref{sec:exp_k_i}.}
  \label{fig:appendix_steps}
\end{figure}

\begin{figure}[htbp]
  \centering
  \includegraphics[width=0.8\textwidth]{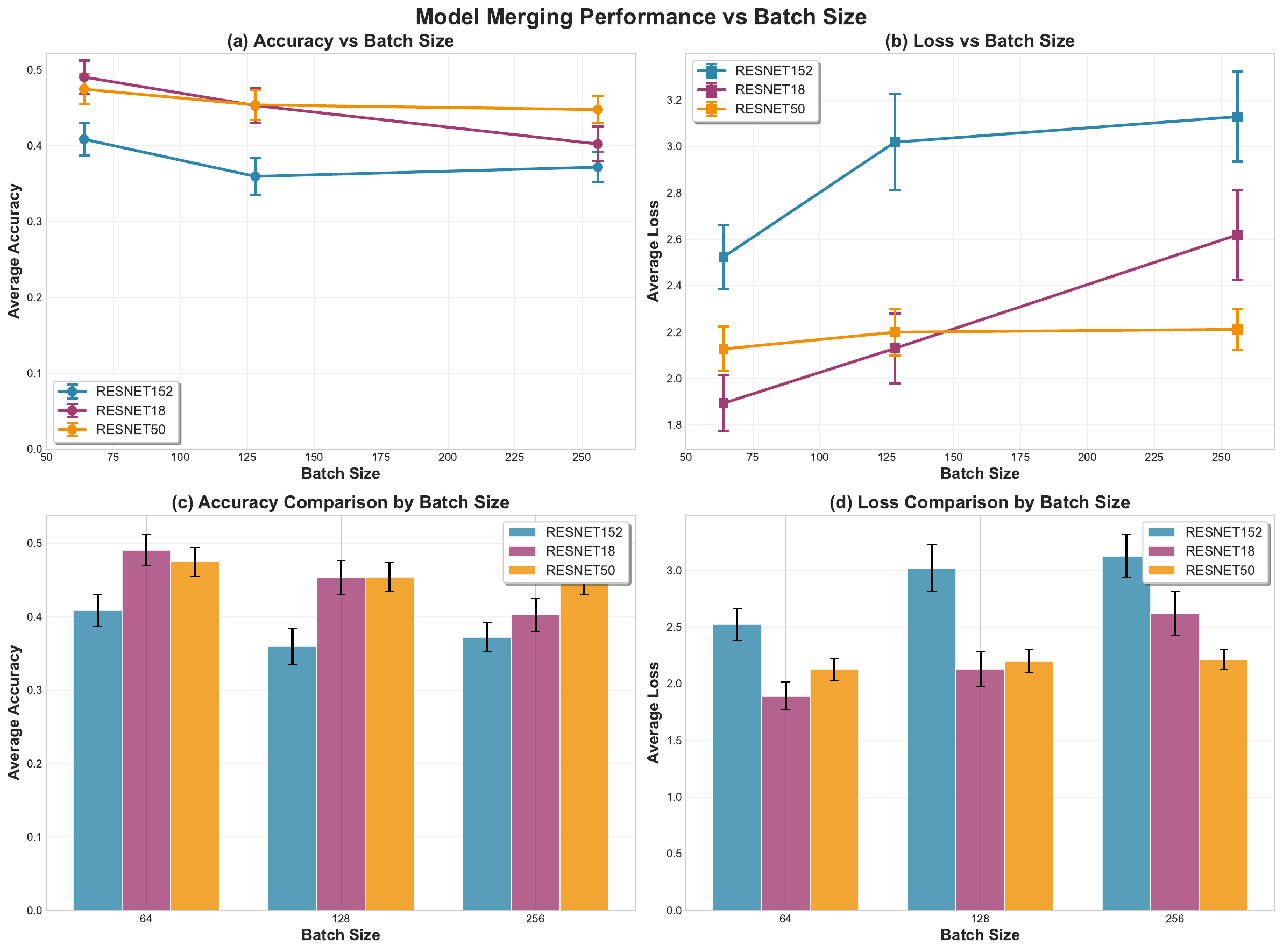} 
  \caption{\textbf{Detailed results for the impact of batch size ($b_i$).} The plots show that for all three backbones, increasing the batch size from 64 to 256 leads to an improvement in performance, as evidenced by decreasing average accuracy (a) and increasing average loss (b). This suggests that in our experimental setting, the potential benefits of reduced gradient variance—a key component of our theoretical bound—outweigh the negative impact on model stability from larger batches, as discussed in Section~\ref{sec:b_i}.}
  \label{fig:appendix_batch_size}
\end{figure}

\begin{figure}[htbp]
  \centering
  \includegraphics[width=0.8\textwidth]{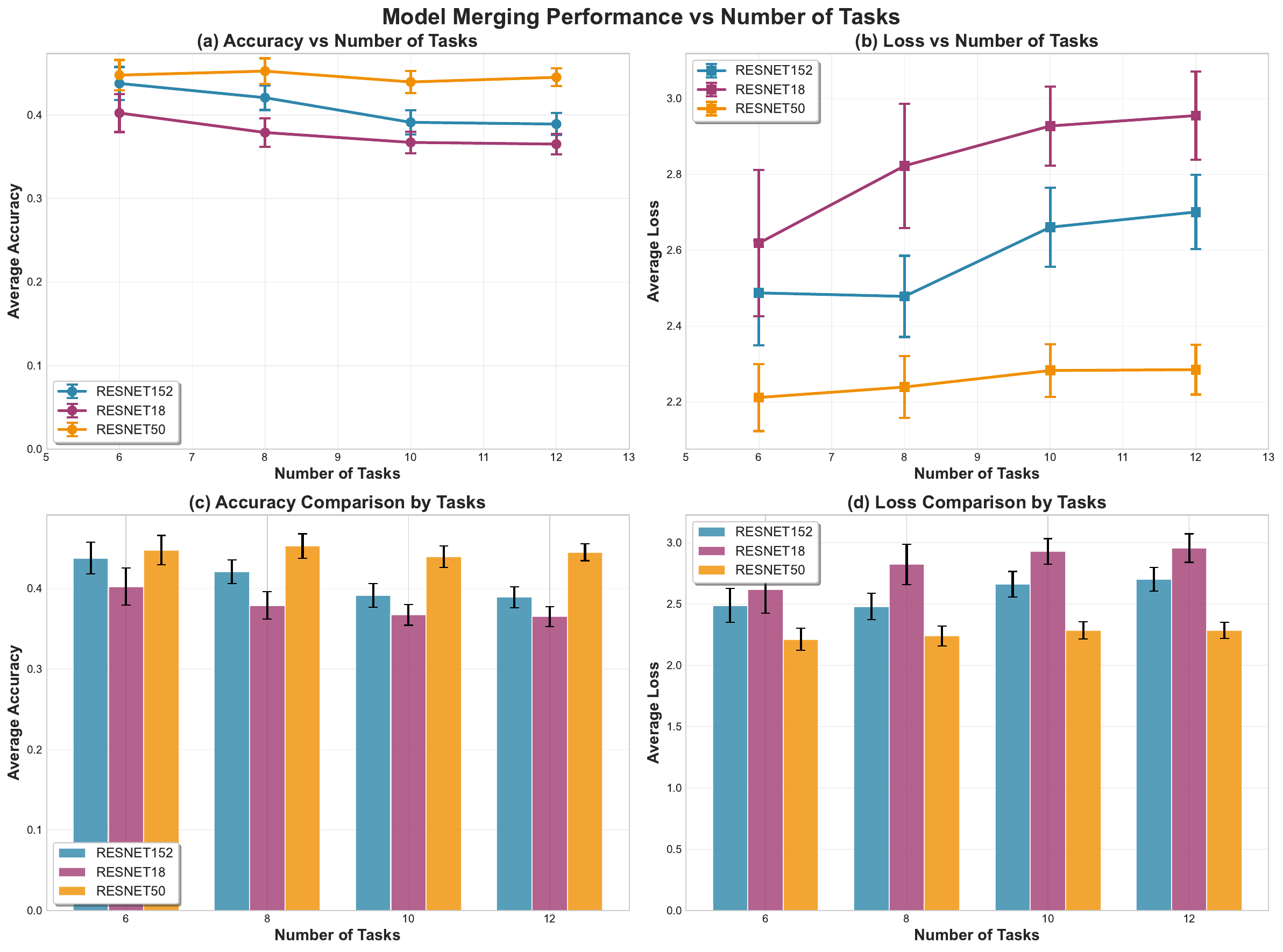} 
  \caption{\textbf{Detailed results for the impact of the number of merged tasks ($N$).} As predicted by our theoretical analysis, increasing the number of tasks leads to a consistent decline in generalization performance. This is shown by the decreasing accuracy trends (a) and increasing loss trends (b) across all ResNet backbones. The results confirm that the penalty from accumulating task heterogeneity is the dominant factor, overwhelming the marginal benefits of optimization, as discussed in Section~\ref{sec:num_of_task}.}
  \label{fig:appendix_num_tasks}
\end{figure}

\begin{figure}[htbp]
  \centering
  \includegraphics[width=0.8\textwidth]{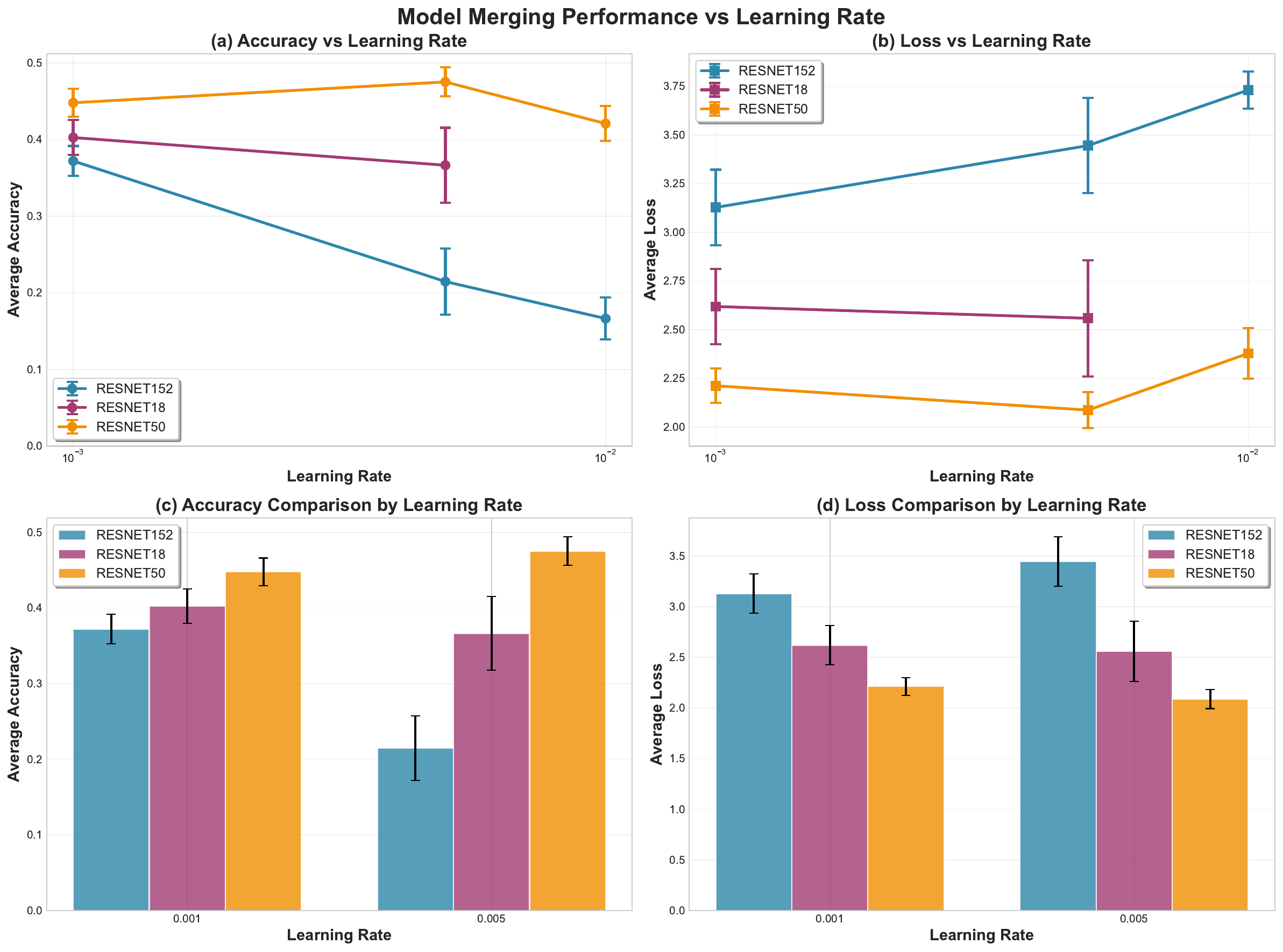} 
  \caption{\textbf{Detailed results for the impact of the learning rate ($\eta_l$).} The plots demonstrate the critical role of the learning rate in model merging. A larger learning rate leads to a severe degradation in performance, with sharply decreasing accuracy (a) and increasing loss (b). This provides strong empirical evidence for our theoretical claim that the learning rate is a dominant factor controlling model stability, with larger values causing the stability component of our error bound to explode, as discussed in Section~\ref{sec:eta}}
  \label{fig:appendix_lr}
\end{figure}

\begin{figure}[htbp]
  \centering
  \includegraphics[width=0.8\textwidth]{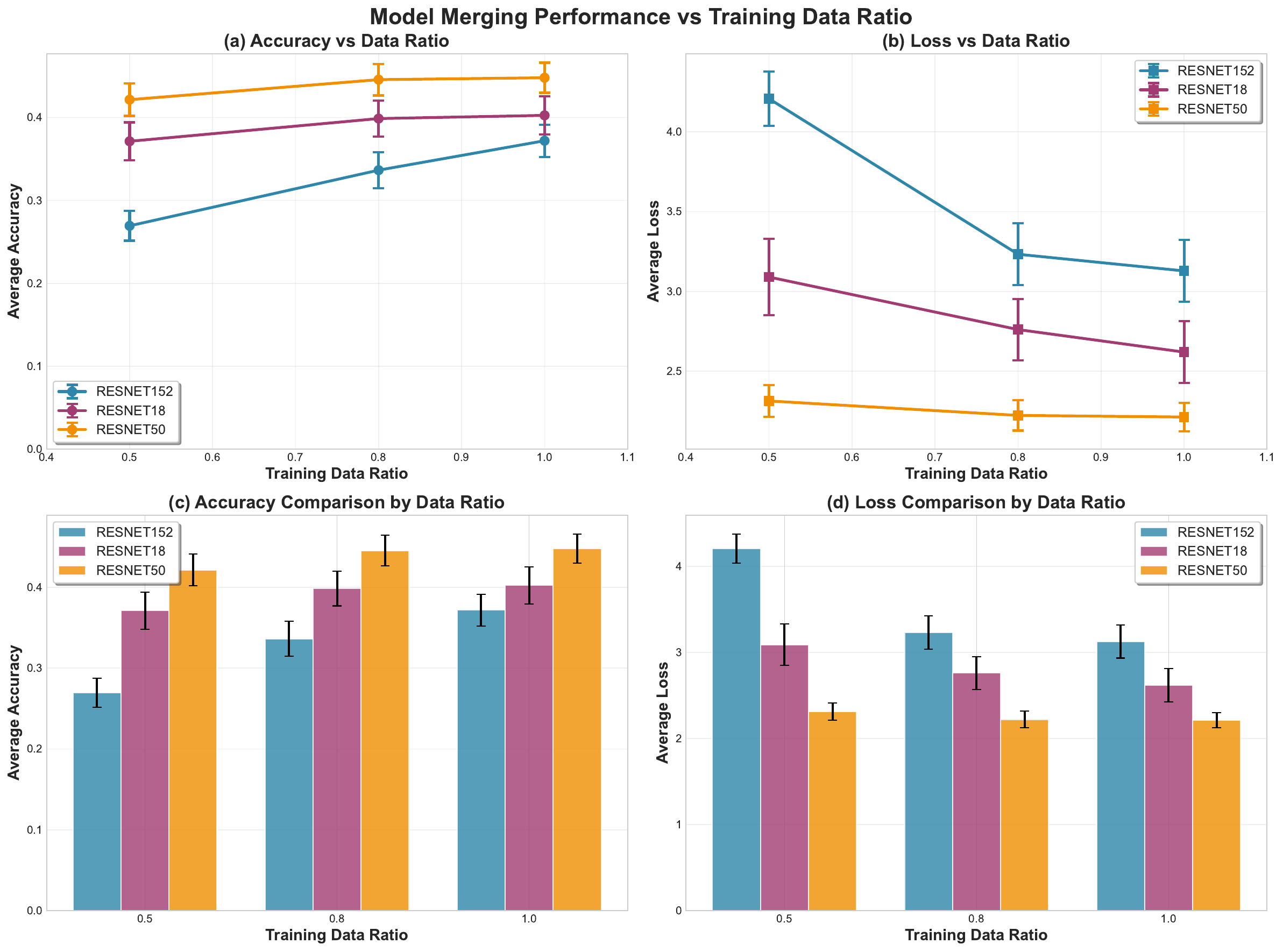} 
  \caption{\textbf{Detailed results for the impact of the training data ratio ($\alpha_i$).} The results show a clear and monotonic improvement in performance as more training data is used. Across all backbones, average accuracy (a) increases and average loss (b) decreases with a larger data ratio. This aligns perfectly with our theoretical prediction that more data enhances model stability and strictly tightens the generalization error bound, providing a "free lunch" for generalization, as discussed in Section~\ref{sec:data_ratio}.}
  \label{fig:appendix_data_ratio}
\end{figure}

\end{document}